\pdfoutput=1
\documentclass[11pt]{article}

\usepackage[preprint]{acl}

\usepackage{times}
\usepackage{latexsym}
\usepackage{graphicx}
\usepackage{subcaption}
\usepackage{amssymb,amsthm,dsfont,mathrsfs}
\usepackage{amsmath}
\usepackage{setspace}
\usepackage{multirow}
\usepackage{helvet}
\usepackage{courier}
\usepackage{bm}
\usepackage{mdwlist}
\usepackage{booktabs}
\usepackage{algorithm}
\usepackage{algorithmic}
\usepackage{url}
\usepackage{colortbl}
\usepackage{enumerate}
\usepackage{enumitem}
\usepackage[textsize=tiny]{todonotes}

\usepackage[T1]{fontenc}
\usepackage[utf8]{inputenc}

\usepackage{microtype}

\usepackage{inconsolata}

\usepackage{xspace}
\newcommand{\model}{\textsc{Position Bias Mitigates Position Bias:}\xspace}

\title{\model Mitigate Position Bias Through Inter-Position Knowledge Distillation }

\author{
    Yifei Wang\textsuperscript{1,2,3}\footnotemark[1], Feng Xiong\textsuperscript{3}\footnotemark[1], Yong Wang\textsuperscript{3}\footnotemark[2]\footnotemark[3], Linjing Li\textsuperscript{1,2}\footnotemark[2]\\ 
    \textbf{Xiangxiang Chu\textsuperscript{3}}, \textbf{Daniel Dajun Zeng\textsuperscript{1,2}} \\
    $^1$MAIS, Institute of Automation, Chinese Academy of Sciences\\
    $^2$School of Artificial Intelligence, University of Chinese Academy of Sciences\\
    $^3$AMAP, Alibaba Group \\
    \texttt{wangyifei2022@ia.ac.cn, wangyong.lz@alibaba-inc.com}  \\
    \href{https://github.com/AMAP-ML/Pos2Distill}{\texttt{https://github.com/AMAP-ML/Pos2Distill}}\\
}
\begin{document}
\maketitle
{
\renewcommand{\thefootnote}{\fnsymbol{footnote}}
\footnotetext[1]{\ Equal contribution. Work done when Yifei's internship at AMAP, Alibaba Group.}
\footnotetext[2]{\ Corresponding author.}
\footnotetext[3]{\ Project lead.}
}
\begin{abstract}
Positional bias (PB), manifesting as non-uniform sensitivity across different contextual locations, significantly impairs long-context comprehension and processing capabilities. 
Previous studies have addressed PB either by modifying the underlying architectures or by employing extensive contextual awareness training. However, the former approach fails to effectively eliminate the substantial performance disparities, while the latter imposes significant data and computational overhead.
To address PB effectively, we introduce \textbf{Pos2Distill}, a position to position knowledge distillation framework.
Pos2Distill transfers the superior capabilities from advantageous positions to less favorable ones, thereby reducing the huge performance gaps. The conceptual principle is to leverage the inherent, position-induced disparity to counteract the PB itself. We identify distinct manifestations of PB under \textbf{\textsc{r}}etrieval and \textbf{\textsc{r}}easoning paradigms, thereby designing two specialized instantiations: \emph{Pos2Distill-R\textsuperscript{1}} and \emph{Pos2Distill-R\textsuperscript{2}} respectively, both grounded in this core principle. By employing our approach, we achieve enhanced uniformity and significant performance gains across all contextual positions in long-context retrieval and reasoning tasks.
Crucially, both specialized systems exhibit strong cross-task generalization mutually, while achieving superior performance on their respective tasks.
\end{abstract}
\section{Introduction}

\begin{center}
    \fcolorbox{white}{white}{\parbox{.9\linewidth}{
    \centerline{\textit{Who tied the bell could be the one to untie it.}}
    % \vspace{-0.05em}
    \rightline{\textit{--- Chinese proverb}}
}}
\end{center}\label{chinese proverb}
Large Language Models (LLMs) are increasingly proficient in handling long contexts, which has been unlocked by key innovations in efficient attention mechanisms \cite{dao2022flashattention, ainslie2023gqa} and length extrapolation techniques \cite{he2024two, chi-etal-2023-dissecting, press2021train}.
These breakthroughs enable LLMs to tackle complex question answering over substantially larger context windows \cite{agarwal2024manyshot}, marking a crucial step towards more capable and versatile natural language systems.

\begin{figure}[t!] 
  \centering
  \includegraphics[width=0.95\linewidth]{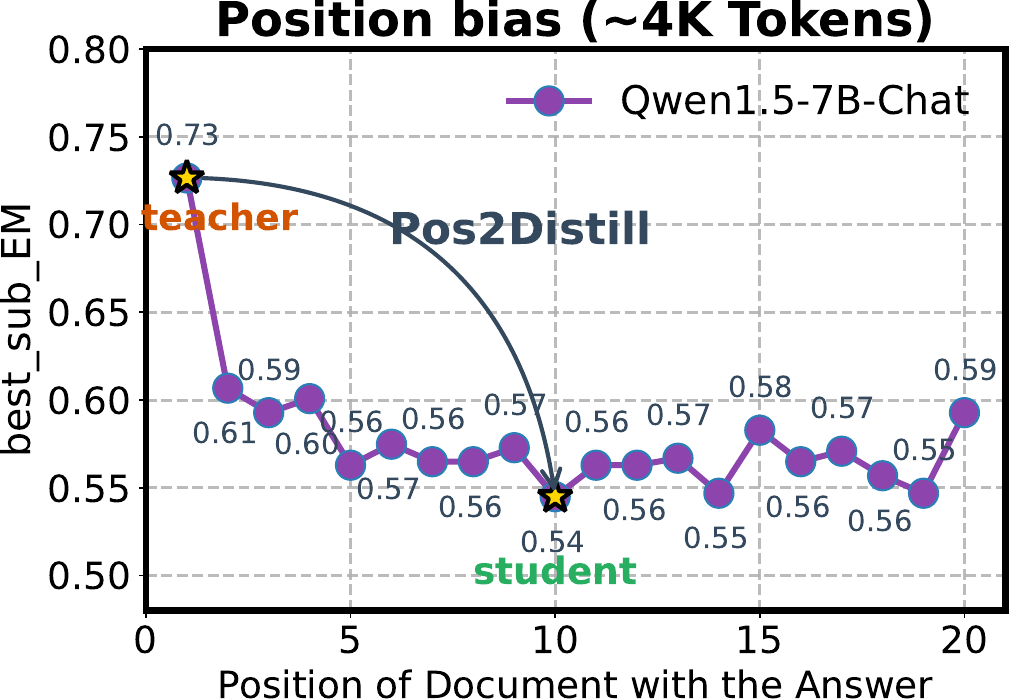}
  \caption{\textbf{Motivation for Pos2Distill}. Marked performance decline from the 1st to 10th position in multi-document QA underscores severe PB. From an alternative viewpoint, superior responses at advantageous positions provide effective supervisory signals for less optimal positions.}
  \label{fig:intro}
  \vskip -0.24in
\end{figure}
Nevertheless, recent studies point out a critical limitation: LLMs do not uniformly extract and utilize information across long context, consistently favoring information located at the context edges while neglecting that in the middle. This phenomenon, commonly termed the \textit{lost in the middle} problem \cite{liu-etal-2024-lost}, underscores a pervasive and intrinsic PB inherent in long-context handling.

PB poses significant obstacles in information-rich settings, such as retrieval-augmented generation~\cite{fang2025attentionrag,dong2025benchmarking}, long-context reasoning \cite{li2024large, kuratov2024babilong}, and LLM-as-a-judge \cite{wang-etal-2024-large-language-models-fair, li-etal-2024-split}. When critical information is distributed arbitrarily throughout the input, LLMs fail to identify and integrate gold content \cite{baker2024lost}, culminating in unexpected model failures across various applications.

To alleviate PB, prior research has delved into its underlying causes,  seeking to modify key architectural components or internal representations linked to uneven contextual sensitivity \cite{zhang2024found, chen-etal-2024-fortify}.
Despite recent progress in narrowing the performance gap, a substantial disparity in information utilization between advantageous and disadvantaged positions still persists. Another line resorts to intensive contextual awareness training \cite{an2024make,zhang2024raft} by synthesizing training data with fine-grained information awareness \cite{zhang2024positionawareparameterefficientfinetuning}. However, such data-driven approaches typically incur substantial costs in both data synthesis and computational resources. Consequently, there remains a critical need for \emph{effective} and \emph{efficient} strategies to mitigate PB that overcome these limitations.

% To alleviate the position bias, prior works have  analyzed its underlying causes and sought to modify key architectural components or internal representations associated with uneven contextual sensitivity. A broad spectrum of interventions has been explored, spanning position encodings \cite{zhang2024found, chen-etal-2024-fortify, lin2024mixture}, causal masks \cite{wang2025eliminating}, as well as internal states such as attention re-weighting \cite{hsieh-etal-2024-found, tan2025pear} and hidden state modulation \cite{yu2024mitigate}. 
% % While these mechanistic approaches incur minimal computational overhead, they are tailored to few tasks or model architectures, thus exhibiting limited generalizability. Furthermore, 
% Despite recent progress in reducing the performance gap, a substantial disparity between advantageous and disadvantaged positions persists.

% These methods construct specialized training data aimed at fine-grained information awareness or enhanced position sensitivity \cite{zhang2024positionawareparameterefficientfinetuning}, providing explicit supervision to identify relevant information throughout the entire context window. However, such data-driven approaches typically incur substantial costs in both data synthesis and computational resources, which significantly limit their scalability.

Inspired by the proverb \ref{chinese proverb}, we contend that PB not only imposes challenges but also implicitly reveals gold signals, which can be exploited to mitigate position-induced disparity (Fig.~\ref{fig:intro}). Our analysis further reveals that PB exhibits distinct behavior under retrieval and reasoning paradigms. In retrieval tasks, PB predominantly manifests as token-shifting, whereas in reasoning tasks, PB interacts with Chain-of-Thought (CoT) processes~\cite{wei2022chain}, leading to thought-shifting, characterized by deviations in the reasoning trajectory.

To this end, we introduce \textbf{Pos2Distill}, a novel position to position knowledge distillation framework, transferring knowledge from advantageous positions to rectify responses at unfavorable ones.
% In retrieval, PB primarily manifests as token-shifting. In contrast, under reasoning mode, PB interacts with Chain-of-Thought (CoT) \cite{wei2022chain} processes, leading to thought-shifting, where the entire reasoning trajectory is skewed.
Customarily, we develop two systems: Pos2Distill-R\textsuperscript{1} and Pos2Distill-R\textsuperscript{2}. Pos2distill-R\textsuperscript{1} mitigates token-shifting in retrieval by incorporating  Kullback-Leibler (KL) divergence loss \cite{Kullback1951KL}, providing fine-grained corrective signals. Pos2Distill-R\textsuperscript{2} addresses thought-shifting in reasoning tasks by distilling high-quality CoT responses from advantageously positioned inputs to guide and rectify reasoning trajectories at less favorable positions.

Extensive experiments demonstrate that Pos2Distill leads to more uniform and substantially improved performance both for in-context retrieval and reasoning tasks. Furthermore, data efficiency is a notable property of our method: with only 250 training instances, Pos2Distill increases the performance of Mistral-7B-v0.3 on the NQ dataset by 6.7\%, as indicated in Fig.~\ref{fig: ablations for pos2-1}.

Our contributions can be summarized as follows: 
\begin{itemize}[leftmargin=*,topsep=1pt,itemsep=0.4pt]
\item We uncover distinct manifestations of PB under retrieval and reasoning paradigms, namely \emph{token-shifting} and \emph{thought-shifting}, offering deeper insights into the nature of PB.

\item We propose \textbf{Pos2Distill}, a novel position to position KD framework to mitigate PB. Given the different behavior of PB, we design two systems tailored to address the PB in both scenarios.

\item Extensive experiments and ablation studies thoroughly demonstrate the efficacy in terms of performance, data efficiency, and superior generalization.
\end{itemize}

\section{Related Work}
\label{sec: related work}

\begin{figure}[t!]
    \centering
    \begin{subfigure}{0.49\linewidth}
        \includegraphics[width=\linewidth]{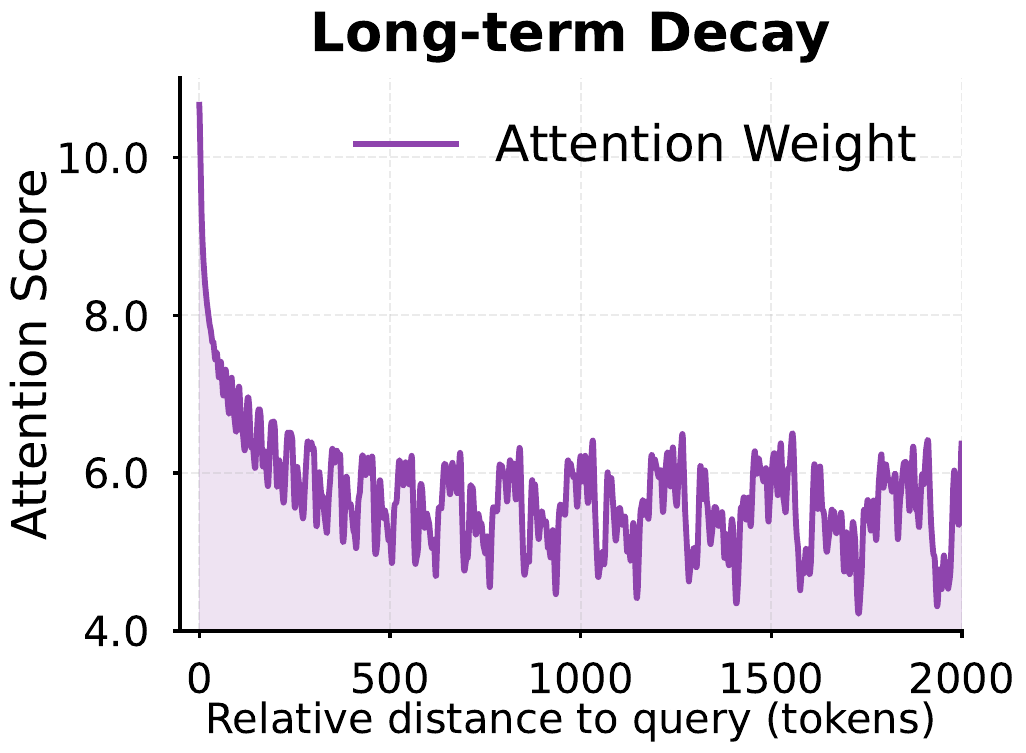}
        \label{longdecay}
    \end{subfigure}
    \hfill
    \begin{subfigure}{0.49\linewidth}
        \includegraphics[width=\linewidth]{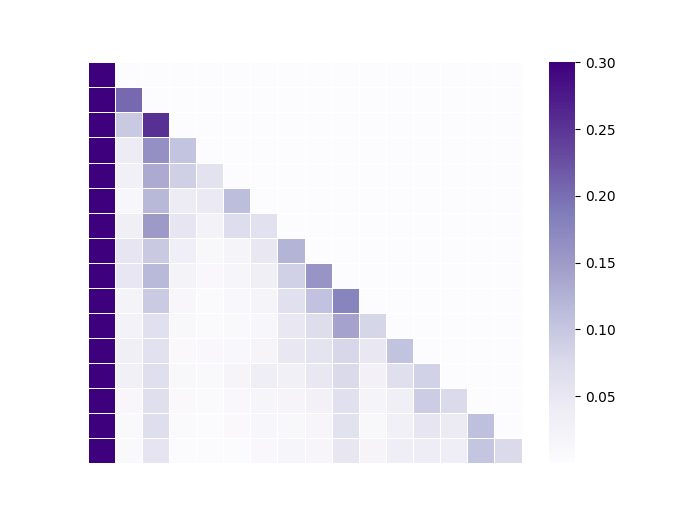}
        \label{attention sink}
    \end{subfigure}
    \vspace{-.9cm}
    \caption{Left: long-term decay effect property of RoPE; Right: Attention sink (causal attention matrix).}
    \vspace{-.1in}
    \label{fig: long-term decay}
\end{figure}

\paragraph{Causes of Position Bias.} 
Most LLMs adopt relative positional encodings \cite{peysakhovich2023attention}, such as RoPE \cite{su2024roformer} and ALiBi  \cite{press2021train},  integrating token relative distances into attention score computation. 
% between the $m$-th and $n$-th tokens. 
This design induces a long-range decay effect in Fig.~\ref{fig: long-term decay} (left), whereby LLMs preferentially attend to recent tokens.
Concurrently, LLMs exhibit a notable bias towards the initial tokens, which can be largely attributed to the universal phenomenon, attention sink \cite{xiao2024efficient,gu2025when}, where disproportionate attention is allocated to early tokens in Fig.~\ref{fig: long-term decay} (right), regardless of semantic significance.
In addition, causal mask enforces a unidirectional flow of information, implicitly encoding positional information \cite{haviv-etal-2022-transformer,chi-etal-2023-latent,wang-etal-2024-length}. The interplay of these factors collectively contributes to the emergence of PB, as further elaborated in Appendix~\ref{appendix related work}.

\paragraph{Mechanistic Approaches.}
Current approaches predominantly focus on aforementioned underlying causes of PB. A broad spectrum of interventions has been investigated, ranging from modifications to position encodings \cite{zhang2024found, chen-etal-2024-fortify, lin2024mixture} and alterations to causal masks \cite{wang2025eliminating}, to the manipulation of internal states, including attention re-weighting \cite{hsieh-etal-2024-found, tan2025pear} and hidden state manipulation \cite{yu2024mitigate}. Despite these efforts, the substantial performance gap across token positions remains largely unmitigated.

  % Differently, \citeauthor{tan2025pear} discover suppression heads which impeding context copying and optimize learnable coefficients to reweight the outputs of the discovered heads. 

% \begin{figure}[t]
%     \centering
%     \begin{subfigure}{0.49\linewidth}
%         \includegraphics[width=\linewidth]{Figures/pos-r1-k-ablation.pdf}
%     \end{subfigure}
%     \hfill
%     \begin{subfigure}{0.49\linewidth}
%         \includegraphics[width=\linewidth]{Figures/pos-r2-k-ablation.pdf}
%     \end{subfigure}
%     \caption{Ablation results for the  number of randomly sampled positions $\mathcal{K}$ for \textsc{Pos2Distill-R\textsuperscript{1}} (a) and \textsc{Pos2Distill-R\textsuperscript{2}} (b).}
%     \label{fig: ablations for pos2-1}
% \end{figure}

\begin{figure}[t]
    \centering
    \begin{subfigure}{\linewidth}
    \includegraphics[width=\linewidth]{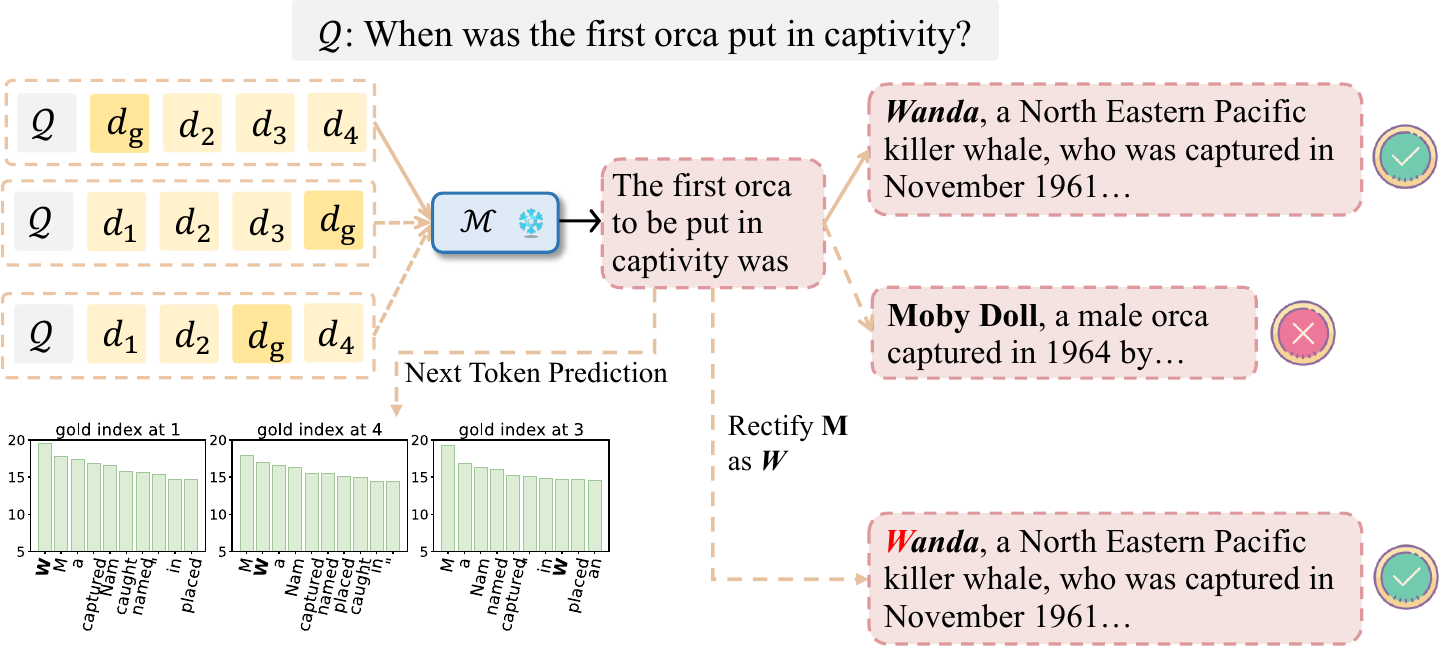}
    \subcaption{Natural PB for Retrieval.
    % Token shifting: PB induces divergence at critical generation steps, resulting in producing erroneous tokens and subsequent retrieval failures.
    % The histogram displays the top-10 logits, with the correct token still ranked highly, illustrating why this phenomenon is termed token shifting in retrieval.
    }
    \label{token-shifting}
    \end{subfigure}
    \vspace{0.1cm}
    \begin{subfigure}{\linewidth}
    \includegraphics[width=\linewidth]{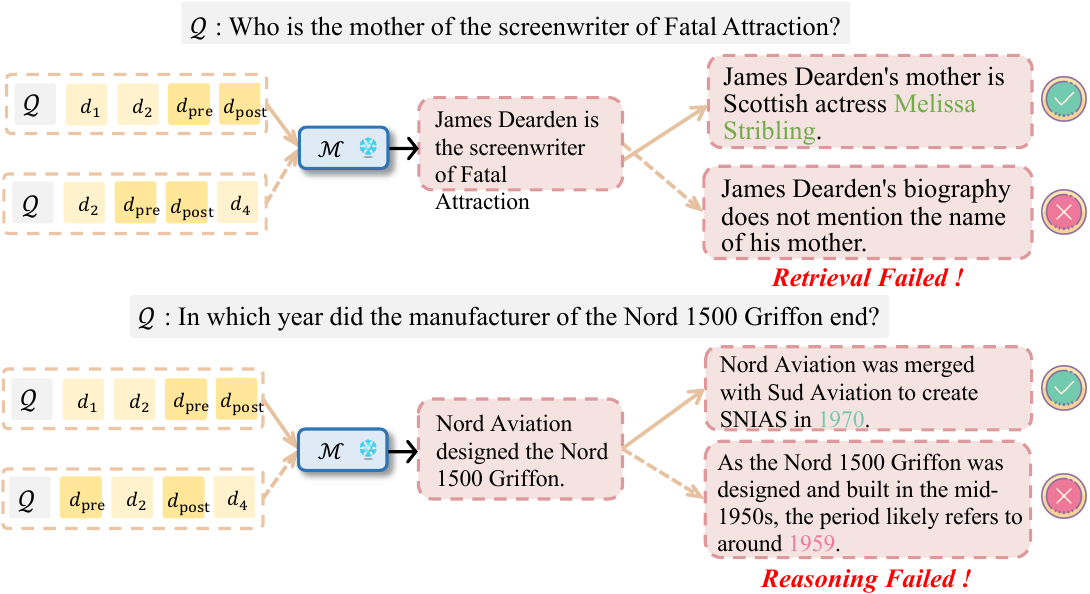}
    \subcaption{
    Compound PB for Reasoning.
    % Thought-shifting: PB interacts with the structure of reasoning chain, could result reasoning failed or retrieval failed.
    }
    \label{thought-shifting}
    \end{subfigure}
    \caption{Different Behavior of Position Bias.}
    \label{fig:motivation}
    \vspace{-0.15in}
\end{figure}

% \begin{figure}[t]
%     \centering
%     \begin{subfigure}{\linewidth}
%     \includegraphics[width=\linewidth]{Figures/Meta-Llama-3-8B-Instruct_ppl.pdf}
%     \end{subfigure}
%     \begin{subfigure}{\linewidth}
%     \includegraphics[width=\linewidth]{Figures/Meta-Llama-3-8B-Instruct.pdf}
%     \end{subfigure}
%     \caption{\textbf{Top}: Average PPL over 500 examples for $\mathcal{R}_{sink},\mathcal{R}_{recent}$ and $ \mathcal{R}_{sft}$ conditioned on $\{\mathcal{P}_i\}_{i=2}^{n-1}$. \textbf{Bottom}: Visualization of the top-10 tokens in the logit distributions at the divergence points. }
%     \label{fig:topk-tokens}
%     \vspace{-0.15in}
% \end{figure}
\paragraph{Training Approaches.} 
Standard next-token prediction pretraining often yields LLMs with inadequate contextual awareness~\cite{shi-etal-2024-replug}. A research direction therefore explores specialized datasets designed to foster fine-grained information awareness~\cite{zhang2024positionawareparameterefficientfinetuning}, training LLMs to identify pertinent information within their full context. Nevertheless, the considerable data and computational overhead limit their scalability and practical viability.

\section{Behavior of Position Bias}\label{sec: Behavior of Position Bias}
\noindent\textbf{Natural PB for Retrieval.}
% \noindent\textbf{Token-shifting of Natural PB for Retrieval.}
Given documents with identical constituent elements arranged in varying orders, our empirical observations indicate that the model's output variability is not uniform across the entire response, but rather highly concentrated at a few decisive turning positions\footnote{Please refer to Appendix~\ref{appendix:behavior_position_bias} for more details.}. 
% Our empirical study reveals that model sensitivity to input variations is not uniformly distributed across the entire output sequence, but rather highly concentrated at a few decisive turning positions\footnote{Please refer to}.
This heterogeneous sensitivity leads to \textbf{token shifting}: the model produced erroneous tokens at these critical positions, consequently retrieval failed. Specifically, as illustrated in Fig.~\ref{token-shifting}, even with consistent prefix output, an erroneous generation of "M" instead of the correct "W" triggers retrieval failure. Moreover, we observe that manual correction of this misaligned token (e.g., changing "M" to "W") enables the model to successfully resume generation and complete the retrieval task. This finding uncovers a \textbf{token recovery} mechanism: once a misaligned token resulting from shifting is rectified, the model can realign its subsequent output. (More supported analysis experiments are provided in Appendix~\ref{appendix: Additional Preliminary Results} .)

\noindent\textbf{Compound PB for Reasoning.}
In-context reasoning fundamentally incorporates two  processes: retrieval and reasoning. The two processes are deeply intertwined, creating a virtuous cycle and producing  numerous outputs~\cite{ren2025efficient,ren2024learn}.  Chain-of-Thought (CoT) reasoning guides the model to formulate more targeted queries, thereby enhancing the precision of the retrieval process \cite{wang-etal-2024-unveiling}.  Conversely, access to accurate, retrieved information provides the factual grounding needed to steer and validate the reasoning steps, thus ensuring the logical integrity of the reasoning chain. At the same time, the selection of reasoning path is very sensitive to irrelevant contexts \cite{yang2025llmreasoningdistractedirrelevant}.
Within this setting, PB is reflected both in variations during retrieval and in alterations that occur during reasoning. Examples of these two types of failures are presented in Fig.~\ref{thought-shifting}. Therefore, it is crucial to reshape the overall response trajectory by integrating genuinely relevant information and reasoning chain.

% In-context reasoning fundamentally comprises retrieval and reasoning. 
% In this context, PB serves as both attention模式的转变 and evolving logical pathway的转变.我们分别称为thought-shifting和retrieval shifting。 As Figure~\ref{thought-shifting}, permuting input documents can cause these two fail. Thus, accurately reconstructing the reasoning trajectory—including pertinent information and logical steps—is essential.

% Comparatively, 当cot被引入时，PB interact the intrinsic structure of the reasoning chain. This interplay 表现为一种compound的印象，not only alters the model's attentional focus but also reshapes the structural configuration of the reasoning logic, sometimes even modifying crucial reasoning steps distributed. We define this transformation within the reasoning chain as \textit{Thought-shifting}, comprising two main type: retrieval failure and reasoning failure.
% As shown in~\ref{thought-shifting}, 在PB的复合作用下，模型可能会经历检索失败和reasoning failue。
% Consequently, it becomes necessary to reconstruct the reasoning process in such instances.
% Thought-shifting: 由偏差导致的推理链路的发生的转变。 我们发现这种转变包括两个方面，分别是检索失效和推理失效。

\section{Methodology} 
PB behaves differently in retrieval and reasoning; thus, a unified approach fails to capture their inherent distinctions. Therefore, we propose two tailored position-aware distillation strategies: (1) For \textbf{R}etrieval (Pos2Distill-R\textsuperscript{1}), we directly calibrate token shifting. (2) For \textbf{R}easoning (Pos2Distill-R\textsuperscript{2}), we transfer high-quality reasoning patterns from advantageous to suboptimal positions, thereby effectively mitigating the compounded bias.
\begin{figure*}[t!]
    \centering
    \begin{subfigure}{0.59\linewidth}
        \includegraphics[width=\linewidth]{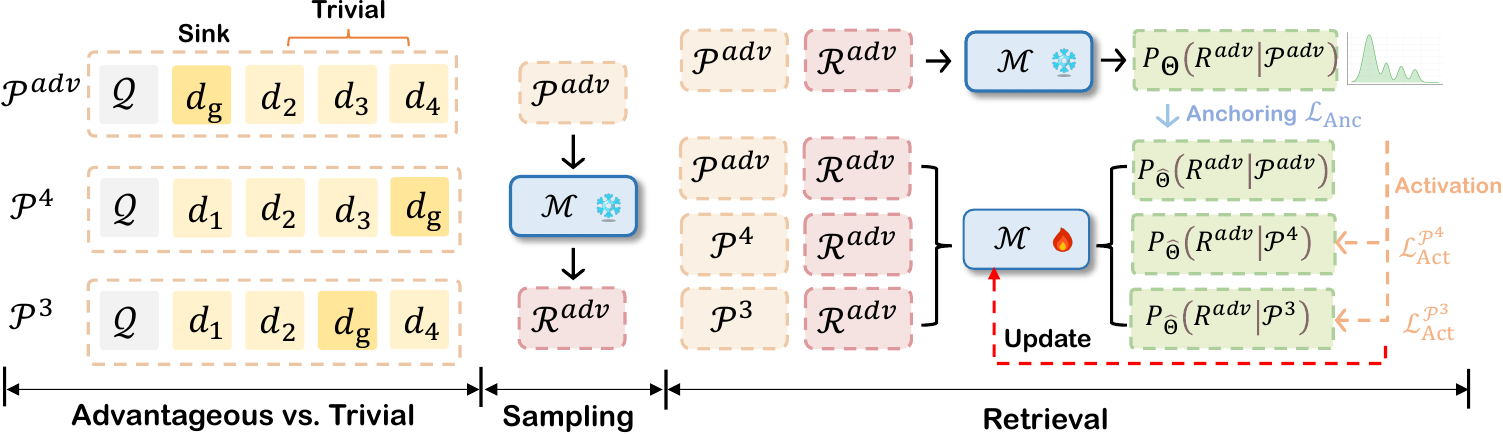}
        \subcaption{\textbf{Pos2Distill-R}\textsuperscript{1} for retrieval.}
        \label{fig:pos2distill_r1}
    \end{subfigure}
    \hfill
    \begin{subfigure}{0.4\linewidth}
        \includegraphics[width=\linewidth]{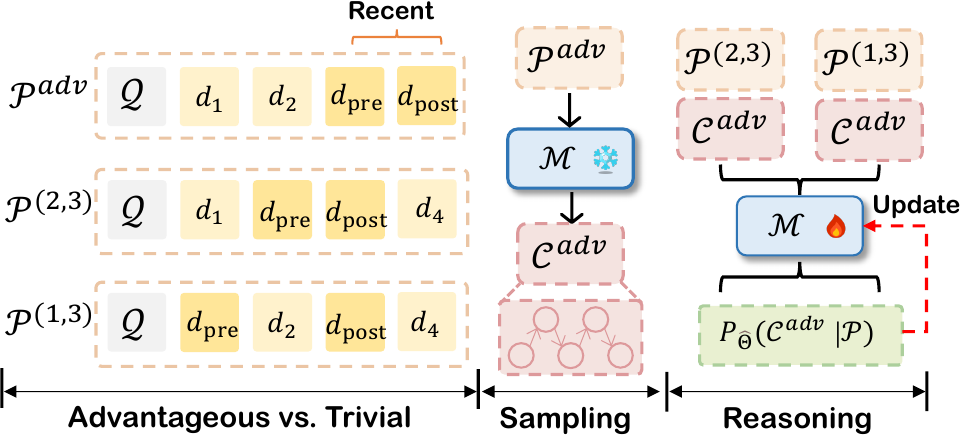}
        \subcaption{\textbf{Pos2Distill-R}\textsuperscript{2} for reasoning.}
        \label{fig:pos2distill_r2}
    \end{subfigure}
    \caption{The design of our \textbf{Pos2Distill}. The left subfigure illustrates \textbf{Pos2Distill-R}\textsuperscript{1}, highlighting its two key strategies: Activation of Trivial Positions and Anchoring of Advantageous Positions. The right subfigure depicts \textbf{Pos2Distill-R}\textsuperscript{2}, designed to reshape reasoning trajectories.}
    \label{fig:method}
    \vspace{-4pt}
\end{figure*}
\subsection{Preliminary}
% \textcolor[RGB]{120,165,212}
% \textcolor[RGB]{252,161,124}
% \definecolor{activation}{RGB}{252,161,124}
% \definecolor{anchoring}{RGB}{120,165,212}

\noindent\textbf{Task Definition.} Following \citet{wang2025eliminating}, we formally define a long-context task as follows. Given the task-specific instruction $\mathcal{I}$, a set of $n$ retrieved documents $\mathcal{D} \hspace{-1mm}:= \{ d_i \}_{i=1}^n$, and a context-dependent question $\mathcal{Q}$, 
a specific LLM $\mathcal{M}$ parameterized by $P_{\Theta}$, generates a response conditioned on the corresponding prompt $\mathcal{P}\hspace{-1mm} := \{ \mathcal{I} \hspace{-1mm}\mid \gamma(\mathcal{D}) \hspace{-1mm}\mid \mathcal{Q} \}$,
where the function $\gamma$ determines the specific ordering of documents in $\mathcal{D}$. 

\noindent\textbf{Retrieval vs. Reasoning.}
\textit{Retrieval} tasks involve identifying an answer directly present in a single document $d_{\text{gold}}\in \mathcal{D}$~\cite{dong2025mmdocir,liu2025beyond}. When $d_{\text{gold}}$
is located at index $i$ within $\mathcal{D}$, the associated  prompt is denoted as $\mathcal{P}^i$, and the response to this prompt is $\mathcal{R}^* \sim \mathcal{M}(\mathcal{P}^*)$. 
In contrast, \textit{reasoning} tasks require $\mathcal{M}$ to integrate information from multiple documents. For formulation brevity, we focus on the two-hop reasoning setting, which involves a first-hop document $d_{\text{pre}}$ at index $i$ and a second-hop document $d_{\text{post}}$ at index $j$  in $\mathcal{D}$. The reasoning prompt is denoted as $\mathcal{P}^{(i,j)}$, and the corresponding response is $\mathcal{C}^* \sim \mathcal{M}(\mathcal{P}^{*})$.

\noindent\textbf{Objective.} The primary objective of Pos2Distill is to enable a model $\mathcal{M}$ such that, for any input $\mathcal{P}^i$ during generation, $\mathcal{M}$ emulates a scenario in which a gold document or a pair of sequentially relevant documents are placed at the most advantageous positions within the input context. 
Specifically, the advantageous position is designated as the ``sink position'' ($\mathcal{P}^1$) for \textit{Retrieval}  and as the ``recent position'' ($\mathcal{P}^{(n-1,n)}$) for  \textit{Reasoning} (shown in Fig.~\ref{fig:method}).

\subsection{Pos2Distill-R\textsuperscript{1} for Retrieval}

% 'Activation of Trivial Positions' seeks to enhance the model's retrieval performance in document positions where it typically underperforms ('trivial positions'). This is accomplished by guiding the model to learn from and emulate its effective processing behaviors observed at 'advantageous positions.' The objective is to enable the model to successfully identify and utilize critical information even when documents are situated in these challenging trivial positions, thereby 'activating' their latent potential.

% Concurrently, 'Anchoring of Advantageous Positions' emphasizes that while striving to improve performance in trivial positions, it is crucial to preserve the model's established high performance and inherent strengths at 'advantageous positions,' ensuring these capabilities are not compromised or degraded.

% Our method proposes a fine-grained knowledge distillation strategy underpinned by two complementary principles: 

\paragraph{Overall Framework.} This section presents Pos2Distill-R\textsuperscript{1}, engineered to calibrate token-shifting behavior and hence mitigate PB for retrieval tasks.
The framework consists of two core modules: \textit{Activation of Trivial Positions} and \textit{Anchoring of Advantageous Positions} in Fig.~\ref{fig:pos2distill_r1}.
The former facilitates the transfer of effective processing capabilities from high-performing advantageous positions to underutilized trivial positions, while the latter ensures the preservation of established performance at advantageous positions, thereby narrowing the gap between trivial and advantageous positions. 

\noindent\textbf{Trivial vs. Advantageous Positions.} For retrieval tasks, we formally define the advantageous position as the first position (sink position), with the remaining positions $\{2,\ldots, n\}$ designated as trivial ones.
For $d_{gold}$ occupies the sink position, the attention sink region overlaps with $d_{gold}$, thereby generating high-quality outputs that inherently imply optimal and robust attention patterns (discussed in ~\ref{sec: related work}) \footnote{Extensive empirical evidence also demonstrates superior performance at the sink position.}.

\noindent\textbf{Activation of Trivial Positions.} To rectify token-shifting behavior, we leverage KL divergence as a fine-grained alignment signal at each generation step.
Concretely, we first sample responses $\mathcal{R}^{\text{adv}} \sim \mathcal{M}(\mathcal{P^{\text{adv}}})$ from advantageous positions. 
We then construct $\mathcal{K}$ prompts, each corresponding to a distinct trivial position, denoted as $\{\mathcal{P}^{n_k}\}_{k=1}^\mathcal{K}$ with $n_i \in \{2,\ldots, n\}$. Our objective is to align the $\mathcal{M}$'s predicted probability distribution over $\mathcal{R}^{\text{adv}}$, conditioned on each trivial prompt $\mathcal{P}^{i}$, with the native distribution over $\mathcal{R}^{\text{adv}}$ conditioned on $\mathcal{P}^{\text{adv}}$.
Formally, for a given position $n_i$,  knowledge distillation at $n_i$ is defined as:
\begin{align*}
\small
\mathcal{L}_{\text{Act}}^{\mathcal{P}^{n_i}} = \mathbb{E}\left[
  \mathrm{KL} \left( P_\Theta(\mathcal{R}^{\text{adv}} | \mathcal{P}^{\text{adv}}) \;\|\;
  P_{\widehat{\Theta}}(\mathcal{R}^{\text{adv}} | \mathcal{P}^{n_i}) \right)
\right],
\end{align*}
where $P_\Theta(\mathcal{R}^{\text{adv}} | \mathcal{P})$ denotes the probability distribution over outputs $\mathcal{R}^{\text{adv}}$ conditioned on input $\mathcal{P}$. Here, $\widehat{\Theta}$ represents the updated parameters at the current step.  Notably, throughout training, we consistently treat $P_\Theta(\mathcal{R}^{\text{adv}} | \mathcal{P})$ as the teacher distribution rather than $P_{\widehat{\Theta}}$, avoiding the loss of advantages at the sink position during model updates.

\textbf{Position-Aware Alignment.}
However, due to the impact on different trivial positions induced by PB varies, the alignment difficulty between $P_\Theta(\mathcal{R}^{\text{adv}})| \mathcal{P}^{\text{adv}})$  and $P_{\widehat{\Theta}}(\mathcal{R}^{\text{adv}} | \mathcal{P}^{n_i})$ is position-dependent. 
Intuitively, positions with higher alignment difficulty should be prioritized with gradient updates to better domain adaptation.  
Motivated by this intuition, we introduce \textit{Position-Aware Alignment}, a dynamic learning strategy that adaptively adjusts learning based on alignment difficulty, ensuring balanced and effective training. 
Technically, given a batch of $\mathcal{B}$ examples $\{\mathcal{Q}_b\}_{b=1}^{|\mathcal{B}|}$ and their corresponding $\mathcal{K}$ trivial prompts $\{\mathcal{P}^{n_k}_{j}\}_{k=1}^{\mathcal{K}}$ for each $\mathcal{Q}_j$,  $\mathcal{K}\cdot \mathcal{B}$ prompts are divided into $n-1$ distinct trivial bins based on their trivial positions. Each bin is represented as $\widehat{\mathcal{B}}_i \hspace{-1mm} = \hspace{-1mm}\{ \mathcal{P}^{i}_{j} \mid   \mathcal{P}^{i}_{j} \in \mathcal{B}, j \in \{1,\ldots,|\mathcal{B}|\}\}$. To model alignment difficulty for each trivial position, we introduce an inter-position weighting scheme for position variations induced by PB while introducing an intra-position weight for differentiating instance variations in the same trivial bin  $\widehat{\mathcal{B}}_i$. Therefore, a dynamic weight $\alpha_{ij}$ for each instance  is defined as follows:
\vspace{-1pt}
\begin{align*}
\alpha_{ij} = \underbrace{\frac{
    \exp\left( \frac{1}{|\widehat{\mathcal{B}}_i|}{\sum_{j=1}^{|\widehat{\mathcal{B}}_i|} \mathcal{L}_{\mathrm{Act}}^{\mathcal{P}^i_j} }\right)
}{\sum\limits_{i=2}^{n}\exp\left( \frac{1}{|\widehat{\mathcal{B}}_i|}{\sum_{j=1}^{|\widehat{\mathcal{B}}_i|} \mathcal{L}_{\mathrm{Act}}^{\mathcal{P}^i_j} }\right)
}}_\textit{inter-pos}\cdot 
\underbrace{\frac{\mathcal{L}_{\text{Act}}^{\mathcal{P}^i_j}}{\max\limits_k\left\{\mathcal{L}_{\text{Act}}^{\mathcal{P}^i_k}\right\}}}_\textit{intra-pos}.
\end{align*}
\vspace{-.2cm}
% where the term $\alpha_{ij}$ consists of two parts: an inter-positional term, which reflects the position-aware variations induced by PB, and an intra-positional term, which captures instance-level differences in the same position bin $\widehat{\mathcal{B}}_i$. 
The activation loss of $\mathcal{B}$ can be formulated as:
\vspace{-.1cm}
\begin{align*}
    \mathcal{L}_{\text{Act}} = \sum_{i}^{n}\sum_{j}^{|\widehat{\mathcal{B}}_i|} \alpha_{ij}\mathcal{L}_{\text{Act}}^{\mathcal{P}^{i}_{j}}.
\end{align*}
\vspace{-.7cm}

\paragraph{Anchoring of Advantageous Positions.}
During the distillation process, $\mathcal{M}$ becomes aware that the gold information possibly appear at any location within the context window,  which can dilute significant attention to the sink position, potentially impairing the overall capability on diverse downstream tasks To prevent this, we introduce an anchoring loss to preserve the effectiveness of the advantageous position:
\begin{align*}
\small
\mathcal{L}_{\text{Anc}}\hspace{-0.5mm}=\hspace{-0.5mm} \mathbb{E}\hspace{-0.5mm}\left[\hspace{-0.5mm}
  \mathrm{KL}\hspace{-0.5mm} \left(\hspace{-0.5mm} P_\Theta(\mathcal{R}^{\text{adv}} | \mathcal{P}^{\text{adv}}) \hspace{-1mm}\;\|\;\hspace{-1mm}
  P_{\widehat{\Theta}}(\mathcal{R}^\text{adv} | \mathcal{P}^{\text{adv}}) \right)\hspace{-0.5mm}
\right]\hspace{-0.5mm}.
\end{align*}

\paragraph{Training Objective.} 
Our approach optimizes a composite loss function that integrates activation loss and anchoring loss, formally:
\vspace{-4pt}
\begin{align*}
\mathcal{L} = \mathcal{L}_{\text{Act}} + \lambda\mathcal{L}_{\text{Anc}},
\end{align*}
% \vspace{-3pt}
where $\lambda$ is a hyperparameter controlling the intensity of the anchoring loss in joint learning.

\subsection{Pos2Distill-R\textsuperscript{2} for In-context Reasoning}

% \paragraph{Overall Framework.}
% 本节将介绍 Pos2Distill-R\textsuperscript{2} 框架，其核心设计理念在于重塑推理轨迹，从而有效缓解推理任务中存在的思维偏移现象。
This section introduces the Pos2Distill-R\textsuperscript{2} to reshape reasoning trajectories. 
The core principle underlying this method is to ensure that the correct reasoning process is consistently activated, irrespective of the positions of relevant documents. 
% In other words, Pos2Distill-R\textsuperscript{2} aims to trigger effective reasoning even when relevant information is not advantageously positioned within the input.

% reshape reasoning trajectories to effectively mitigate the issue of reasoning bias commonly observed in inference tasks.
% This section presents Pos2Distill-R\textsuperscript{2}, which is designed to reshape thought-shifting and hence mitigate PB in reasoning tasks. Our goal is to ensure that, regardless of where the relevant documents are positioned, the correct reasoning process is triggered \ywang{as when} they are placed in advantageous positions. 

\paragraph{Trivial vs. Advantageous Positions.}  Although  PB in multi-hop reasoning scenario is sophisticated, insights from \cite{baker2024lostmiddleinbetweenenhancing} allow us to summarize several key patterns: PB is closely associated with the absolute position within the context window, the distance between relevant documents, and their relative order.\footnote{Please refer to Table~\ref{tab: pb for reasoning}.}  
Empirically, placing $d_{\text{pre}}$ and $d_{\text{post}}$ at indices $n-1$ and $n$ is expected to yield optimal performance. Consequently, we formally define the $\mathcal{P}^{\text{adv}}$ for reasoning tasks as $\mathcal{P}^{(n-1,n)}$.

\paragraph{Reshaping Reasoning Trajectories.}
We begin by sampling CoT reasoning trajectories from the advantageous positions $\mathcal{P}^{\text{adv}}$, denoted as $\mathcal{C}^{\text{adv}}\sim \mathcal{M}(\mathcal{P}^{\text{adv}})$.
Similar to the procedure for \textit{Retrieval} tasks, we construct $\mathcal{K}$ distinct prompts for each position set $\{n_k^{pre},n_k^{post}\}$, denoted as $\{\mathcal{P}^{\left(n_k^{\text{pre}},n_k^{\text{post}}\right)}\}_{k=1}^\mathcal{K}$, where $n_k^{\text{pre}}$ and $n_k^{\text{post}}$ are selected from the set $\{1, \ldots, n\}$ subject to $n_k^{\text{pre}} \neq n_k^{\text{post}}$. 
The prompts $\mathcal{P}^{\left(n_k^{\text{pre}},n_k^{\text{post}}\right)}$ and their corresponding reasoning trajectory $\mathcal{C}^\text{adv}$ are subsequently optimized using the cross-entropy (CE) loss function to effectively capture the reasoning patterns. Formally,
% The prompts $\mathcal{P}^{\left(n_k^{\text{pre}},n_k^{\text{post}}\right)}$ and corresponding reasoning trajectory $\mathcal{C}^{\text{adv}}$ are then trained using the cross-entropy loss function to capture the reasoning patterns effectively. 
% The training objective is expressed as,
\vspace{-.5cm}
\begin{align*}
\mathcal{L} = - \sum_{k=1}^{\mathcal{K}}\log \mathcal{M}\left(\mathcal{C}^{\text{adv}} \mid\mathcal{P}^{\left(n_k^{\text{pre}},n_k^{\text{post}}\right)}\right).
\end{align*}

\begin{table*}[!ht]
    % \small
    \centering
    \setlength{\tabcolsep}{1mm}
    \renewcommand\arraystretch{.8}
    \vspace{-2ex}
% 0.8\height
    \resizebox{2.1\columnwidth}{!}{
    \begin{tabular}{l|cccccc|cccccc|cccccc|cccccc}
    \toprule
        \multirow{2}{*}{Methods} &  \multicolumn{6}{@{}c|}{{\bf NQ}} &  \multicolumn{6}{@{}c|}{{\bf KV Retrieval}} &  \multicolumn{6}{@{}c|}{{\bf TQA}} &  \multicolumn{6}{@{}c}{{\bf WebQ}} \\
        & 1st & 5th & 10th & 15th & 20th & Avg. & 0\% & 25\% & 50\% & 75\% & 100\% & Avg. & 1st & 5th & 10th & 15th & 20th & Avg.& 1st & 5th & 10th & 15th & 20th & Avg.\\ 
    \midrule

%     \midrule

%     Qwen2.5-7B &52.1&45.3&44.3&40.7&52.3&46.9&95.6&74.2&64.2&58.8&18.2&62.2\\
%     w/Ms-PoE &\textbf{70.0}&\textbf{51.4}&46.8&42.8&47.0&51.6&91.8 & 59.4 & 71.6 & \textbf{74.4} & \textbf{48.8} & 69.2\\
%     w/SFT &67.4&\textbf{51.4}&\textbf{47.6}&\textbf{48.8}&\textbf{48.0}&  \textbf{52.7} &\textbf{97.2}&\textbf{83.4}&\textbf{80.8}&68.8&35.4&\textbf{73.1} \\
%     w/Pos2Distill &67.4&\textbf{51.4}&\textbf{47.6}&\textbf{48.8}&\textbf{48.0}&  \textbf{52.7} &\textbf{97.2}&\textbf{83.4}&\textbf{80.8}&68.8&35.4&\textbf{73.1} \\
%     \midrule

\textsc{Mistral-7B} & \textbf{72.5}&62.3&60.7&63.9&64.9&64.8&99.8&97.6&62.0&35.6&78.0&74.6&85.2&81.8&81.8&82.6&82.2&82.7&82.7&69.7&64.5&62.4&62.5&68.4\\

+Ms-PoE &67.3&58.7&56.7&60.1&61.5&60.9&99.8&\textbf{97.7}&78.4&75.2&95.3&89.3&81.3&79.3&76.7&79.2&\textbf{83.4}&80.0&76.7&64.1&62.3&59.3&63.5&65.2\\
+SFT&68.3&64.5&66.5&66.7&62.7&65.7&\textbf{100.0}&89.0&89.2&75.6&81.8&87.1&78.2&78.0&76.8&77.2&76.2&77.3&52.7&51.1&52.3&50.7&53.9&52.1\\

+SeqKD
&63.3&59.1&59.3&60.5&57.5&59.9&100.0&86.4&\textbf{93.0}&87.0&93.0&91.8 &77.4&77.6&76.4&77.6&75.8&77.0&57.7&58.9&55.7&56.9&56.1&57.0\\

\rowcolor[rgb]{ .867, .922, .969}+Pos2Distill
&70.5 & \textbf{70.7}& \textbf{71.3}& \textbf{71.9} & \textbf{73.3}& \textbf{71.1}&99.0&95.4&92.6&\textbf{90.0}&\textbf{96.8}&\textbf{94.8} &\textbf{85.3}&\textbf{82.4}&\textbf{84.6}&\textbf{83.4}&81.8&\textbf{83.5}&\textbf{70.3}&\textbf{67.9}&\textbf{68.3}&\textbf{70.5}&\textbf{67.5}&\textbf{68.9}\\
\midrule
% \textsc{Qwen1.5-4B }&57.5&46.5&44.3&46.7&49.5& 48.9&99.6&75.6&64.1&72.3&65.3&75.3&73.1&64.9&65.9&65.5&67.3&67.3&57.8&50.7&52.8&50.5&49.9&52.3\\
% % \textsc{+Ms-PoE}  \\
% \textsc{+SFT}&51.9&51.9&53.7&49.9&51.1&51.7&  98.2&81.4&67.0&69.3&67.1&76.6&72.6&68.1&70.9&70.3&68.5&70.1&49.7&46.3&46.5&46.3&45.3&46.8\\
% \textsc{+Seq\_KD}&49.7&50.1&54.1&55.1&50.5&51.9&100.0&80.2&70.0&73.7&65.7&77.9&72.7&68.1&70.5&71.3&70.5&70.6&50.7&49.7&50.7&49.7&48.5&49.9\\
% \rowcolor[rgb]{ .867, .922, .969}\textsc{+Pos2Distill}&51.7&47.1&52.1&51.7&53.1&51.1&98.6&76.2&	72.5&69.7&71.9&77.8&73.3&66.7&70.5&72.7&69.1&70.4&55.9&53.1&55.1&52.7&53.1&\textbf{54.0}\\

\textsc{Qwen1.5-7B }& \textbf{73.6}&57.3&56.9&57.3&60.9&61.2&\textbf{100.0}&	83.8&38.7	&23.3&	30.3& 55.2&82.4&74.7&73.9&75.4&76.4&76.6&64.3&50.9&51.5&50.3&55.2&54.4\\
 % \textsc{+Ms-PoE}  \\
% w/SPHS
% &67.4&58.7&58.3&62.1&66.1&62.5&97.2 &80.5& 51.5& 37.4& 68.1& 66.9 \\
     +Ms-PoE &67.4 &54.8& 54.2& 57.4& 61.3&59.0&97.4&76.5&4.7&6.3&3.2&37.6&76.4&75.2&69.4&67.4&75.1&72.7&\textbf{65.2}&53.4&54.5&49.7&55.6&55.7\\ +SFT&63.5&59.7&62.5&62.5&62.9&62.2&100.0&97.0	&95.8&88.8&91.2&94.6&77.6&77.0&78.4&77.4&77.4&77.6&54.9&52.5&51.9&52.3&52.5&52.8 \\
     +SeqKD&63.9&58.5&62.5&61.1&57.7&60.7&100.0&	91.6&66.9&	50.5&54.5&72.7&80.2&77.0&80.2&78.8&75.6&78.4&56.1&54.5&55.3&54.5&54.9&55.1\\
     \rowcolor[rgb]{ .867, .922, .969}+Pos2Distill &69.9&\textbf{67.3}&\textbf{68.1}&\textbf{69.1}&\textbf{67.5}&\textbf{68.4}&99.8&\textbf{97.3}&\textbf{96.5}&\textbf{97.5}&\textbf{93.2}&\textbf{96.9}&\textbf{82.6}&\textbf{79.8}&\textbf{79.0}&\textbf{80.6}&\textbf{78.1}&\textbf{80.0}&\textbf{64.9}&\textbf{61.5}&\textbf{61.5}&\textbf{61.3}&\textbf{61.8}&\textbf{62.2}\\
\midrule
% \textsc{Qwen2.5-7B }&58.1&52.5&52.9&54.3&57.9&55.1&96.2&98.2&92.4&96.4&94.8&95.6&79.0&77.2&77.6&76.2&77.8&77.6&46.1&42.1&44.1&43.1&45.3&44.1\\
%  \textsc{+Ms-PoE}  \\
%  \textsc{+SFT} \\
%  \textsc{+Seq\_KD}&\\
%  \rowcolor[rgb]{ .867, .922, .969}\textsc{+Pos2Distill} &\\

% \midrule

    \textsc{Llama-3-8B} & \textbf{67.9}&56.7&53.7&57.9&60.8&59.4  &98.0&85.4 &70.3&83.2&68.5&81.1&85.6&83.4&82.2&84.0&83.2&83.7&\textbf{57.9}&51.8&50.7&50.7&52.3&52.8\\
 +Ms-PoE & 65.7&58.5&57.3&58.2&62.5&60.4&97.5&87.3&78.3&81.2&73.4&83.5&\textbf{86.4}&84.2&81.2&81.5&82.7&83.2&56.2&52.3&52.1&49.8&52.3&52.5\\
    +SFT & 65.1&60.7&62.1&65.3&66.7&63.9&98.6	&93.0&	97.0&	98.0&96.8&96.7&82.8&83.4&\textbf{84.8}&83.2&83.2&83.5&56.8&54.7&55.1&55.3&54.5&55.3\\
   +SeqKD &61.7&58.7&58.9&59.9&61.9 &60.2&\textbf{100.0}&	95.6&	95.2&	\textbf{98.2}
&97.6&97.3&82.4&\textbf{84.6}&83.6&\textbf{84.2}&82.6&83.5&54.5&53.7&52.9&51.9&52.5&53.1\\
    \rowcolor[rgb]{ .867, .922, .969}+Pos2Distill&67.7&\textbf{64.1}&\textbf{68.3}&\textbf{66.7}&\textbf{68.1}&\textbf{67.0}&	98.8&\textbf{96.2}&\textbf{98.2}&	97.0&\textbf{97.8}&\textbf{97.6}&85.6&84.2&84.0&84.1&\textbf{83.8}&\textbf{84.3}&57.7&\textbf{56.3}&\textbf{57.1}&\textbf{56.3}&\textbf{56.2}&\textbf{56.7}\\
    \bottomrule
    \end{tabular}
    }

        \caption{Main results of Pos2Distill-R\textsuperscript{1} on both Retrieval-Augmented QA datasets and KV retrieval (140 KV pairs).}
        \vspace{-3pt}
        \label{tab:main resuts pos1}
\end{table*}

\begin{table}[t]
    \centering
    \small
    \renewcommand\arraystretch{.6}
    \tabcolsep 1.3pt
    \resizebox{\columnwidth}{!}{
    \begin{tabular}{p{0.65cm}lccccccccc}
    \toprule
    &&\multicolumn{5}{c}{\textbf{Retrieval-Augmented QA}} \\
    \cmidrule(r){3-7}
    Num.&Meth.&0\%&25\%&50\%&75\%&100\%&\textbf{Avg.$\uparrow$}&\textbf{GAP.$\downarrow$}&\textbf{LEN.}\\
    \midrule
    \multirow{2}{*}{\textbf{20}}& \textsc{base.}
                     &72.3&	63.3&	61.2&	63.5&	65.1&65.1&11.1&\multirow{2}{*}{\text{3.3k}}\\
    & ours&72.3 &69.5	&67.5	&68.5	&69.7&\textbf{69.5}&\textbf{4.8}\\
    % \hdashline
    \midrule
    \multirow{2}{*}{\textbf{30}}& \textsc{base.}&73.7&	59.5&60.5&61.3&64.1&63.8&14.2&\multirow{2}{*}{\text{4k}}\\
    & ours. &70.3&	67.9&69.3&	67.3&	70.9&\textbf{69.1}&\textbf{3.6}
\\
    \midrule
    \multirow{2}{*}{\textbf{40}}& \textsc{base.}&72.7&		60.1&		61.3&		61.7	&	65.7&64.3&12.6&\multirow{2}{*}{\text{6k}}\\
    & ours.&67.1&		66.7&	68.3&		66.5&	68.3&\textbf{67.4}&\textbf{1.8}
\\
    \midrule
    \multirow{2}{*}{\textbf{50}}& \textsc{base.}&74.7		&56.9&58.&59.5& 66.3&63.1&17.8&\multirow{2}{*}{\text{8k}}\\
    & ours. &69.1&		66.1&	66.7&	66.1&67.9&\textbf{67.2}&\textbf{3.0}
\\
    \bottomrule
    \end{tabular}}
    \caption{Generalization of Pos2Distill-R\textsuperscript{1} on longer context. (Mistral-7B-v0.3 trained on 20 docs).}
    
    \label{tab:posr1 for generation}
    \vspace{-7pt}
\end{table}

\section{Experiments}

% \begin{figure}[t!]
%     \centering
%     \begin{subfigure}{\linewidth}
%         \includegraphics[width=\linewidth]{figures/one_model_iter012_category_bar.pdf}
%         \subcaption{Mistral-7B}
%     \end{subfigure}
%     \hfill
%     \begin{subfigure}{\linewidth}
%         \includegraphics[width=\linewidth]{figures/three_model_iter0_category_bar.pdf}
%         \subcaption{Qwen1.5-7B}
%     \end{subfigure}
%     \caption{Ablation results for \textsc{Pos2Distill-R\textsuperscript{1}}. Each point represents the average accuracy when the gold doc is placed at the 1st, 5th, 10th, 15th, and 20th element, reflecting the  contextual retrieval capability under the existence of natural PB. We increase the number of questions and keep sampling 4 random positions.}
%     \label{fig: ablations for pos2-1}
% \end{figure}

\subsection{Experimental Setup} 

\paragraph{Setup for Pos2Distill-R\textsuperscript{1}.} We apply  Pos2Distill-R\textsuperscript{1} to three LLMs, including Mistral-7B-Instruct-v0.3, Qwen1.5-7B-Chat, and Llama-3-8B-Instruct, all of which exhibit severe PB in retrieval tasks. 
The evaluation leverages three datasets: NaturalQuestions (NQ) \cite{kwiatkowski-etal-2019-natural}, TriviaQA (TQA) \cite{joshi-etal-2017-triviaqa} and WebQA (WebQ) \cite{berant2013semantic}, and a specilized task KV Retrieval \cite{liu-etal-2024-lost}. These datasets are setting as retrieval-augmented QA, with each question accompanied by 20 documents.
\begin{itemize}
[leftmargin=*,itemsep=1.5pt,topsep=0pt,parsep=0pt]
\item \textbf{Baselines:} We compare our approach with base model, Ms-PoE~\cite{zhang2024found}, vanilla SFT and SeqKD \cite{kim2016sequencelevelknowledgedistillation}. 
\item \textbf{Evaluation:} We assess PB by measuring task performance as $d_{\text{gold}}$ is systematically placed at various positions. 
\end{itemize}

\paragraph{Setup for Pos2Distill-R\textsuperscript{2}.} For in-context reasoning, we utilize two capable LLMs: Llama-3.1-8B-Instruct and Qwen2.5-7B-Instruct. The evaluation is performed on  three long-context multi-hop reasoning datasets: Hotpot QA \cite{yang-etal-2018-hotpotqa}, MusiQue \cite{trivedi-etal-2022-musique} and 2WikiMultiHopQA \cite{ho-etal-2020-constructing}. (All details in Appx.~\ref{appendix: Experiments Details.})

\begin{itemize}
[leftmargin=*,itemsep=1.5pt,topsep=1pt,parsep=0pt]
\item \textbf{Baselines:} We compare two self-training baselines utilizing CoT data: \textsc{sealong} ~\cite{li2024large} and \textsc{longfaith-sft} and \textsc{longfaith-dpo}~\cite{yang2025longfaithenhancinglongcontextreasoning}\footnote{See more illustrations about baselines in Appendix~\ref{appendix: Experiments Details.} }. 
\item \textbf{Evaluation:} We evaluate from two aspects: (1) performance gains in long-context reasoning, and (2) performance gap induced by the positions.
\end{itemize}

\subsection{Main Results for Pos2Distill-R\textsuperscript{1}}
\paragraph{Pos2Distill-R\textsuperscript{1} obviously mitigates PB.} 
Tab.~\ref{tab:main resuts pos1} summarizes the performance of various methods across different benchmarks. Our analysis yields two key findings:
First, Pos2Distill-R\textsuperscript{1} demonstrates robust and uniform performance irrespective of the position of $d_{\text{gold}}$, markedly reducing position-induced performance disparities.  For example, on WebQ dataset, Pos2Distill-R\textsuperscript{1} enables Llama-3-8B to achieve an average accuracy of 56.7\% across 20 positions. This performance, comparable to 57.9\%  attained when $d_{\text{gold}}$ is situated at an optimal sink position, illustrates successful knowledge transfer from advantageous to unfavorable positions, the core principle of Pos2Distill-R\textsuperscript{1}. Notably, our method significantly outperforms standard SFT, which suggests that the closer distributions of $\mathcal{R}^{adv}$ and $\mathcal{R}^{trivial}$ allows LLMs to adapt more readily than when learning from markedly different distributions, leading to the high data efficiency detailed in Sec.~\ref{data efficiency}.
Furthermore, Pos2Distill-R\textsuperscript{1} effectively generalizes to longer contexts. When evaluated with contexts containing 20 to 50 documents, Mistral-7B-v0.3 trained on 20-document contexts maintains both high overall accuracy and positional uniformity in Tab.~\ref{tab:posr1 for generation}. Crucially, it exhibits significantly narrowed performance gaps across positions.

\begin{figure}[t!]
\centering
\begin{subfigure}{0.45\linewidth}
    \includegraphics[width=\linewidth]{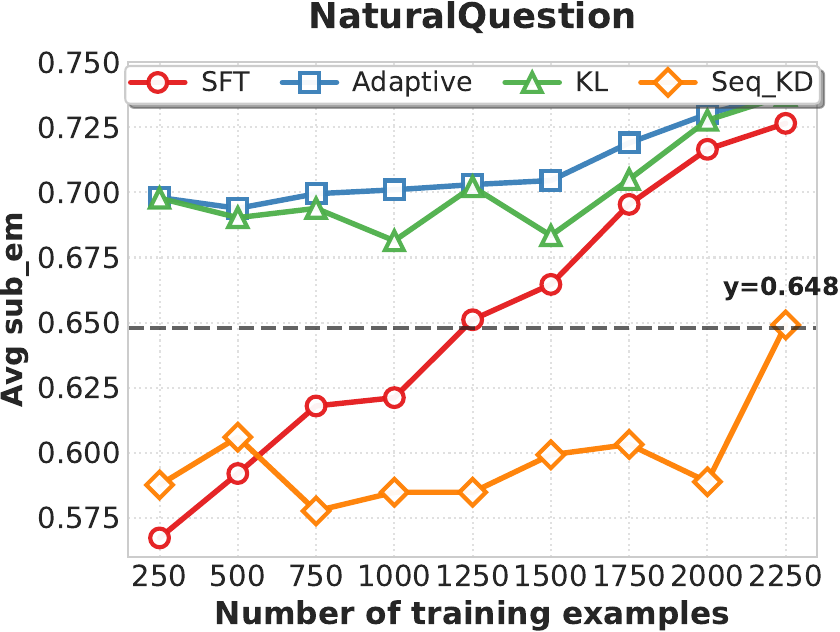}
    % \subcaption{Mistral-7B}
\end{subfigure}
\hspace{-2pt}
\begin{subfigure}{0.45\linewidth}
    \includegraphics[width=\linewidth]{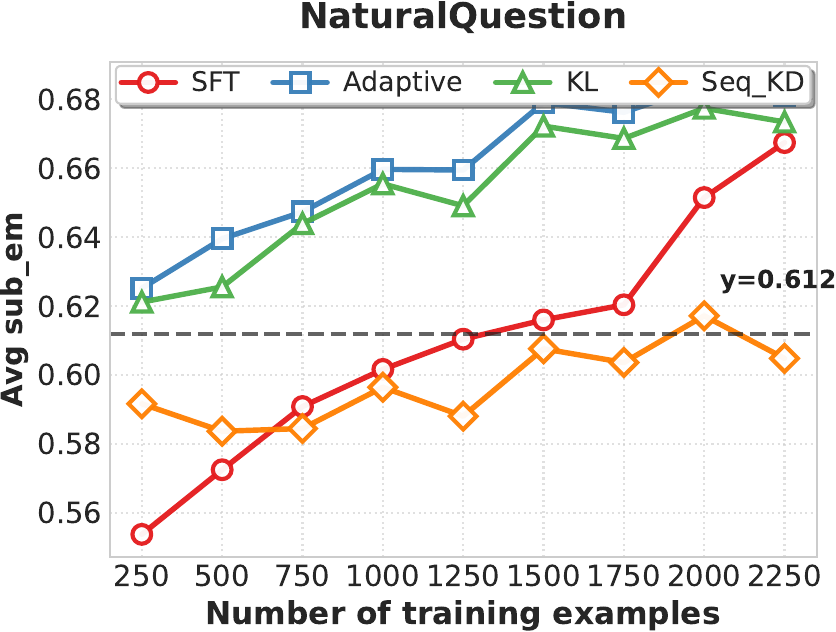}
    % \subcaption{Qwen1.5-7B}
\end{subfigure}
\begin{subfigure}{0.45\linewidth}
    \includegraphics[width=\linewidth]{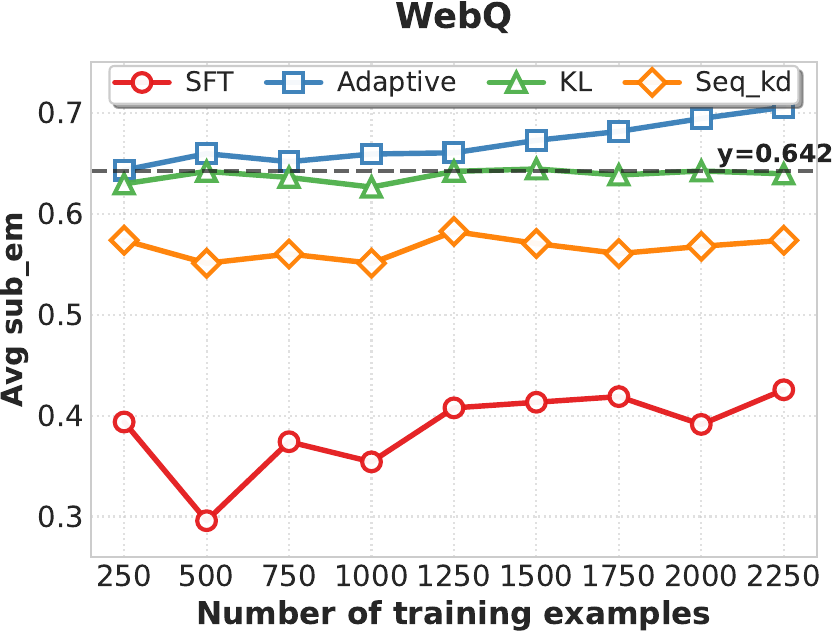}
    \subcaption{Mistral-7B}
\end{subfigure}
\hspace{-1pt}
\begin{subfigure}{0.45\linewidth}
    \includegraphics[width=\linewidth]{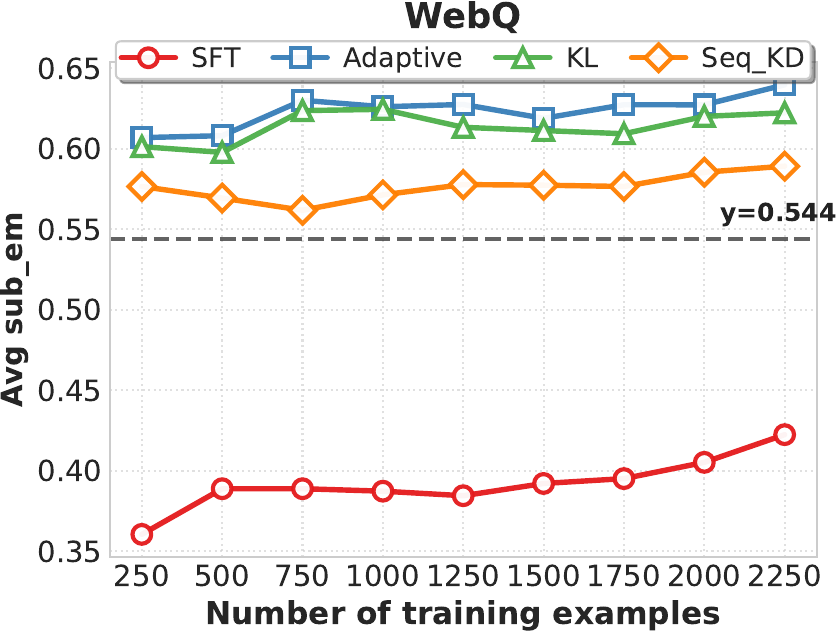}
    \subcaption{Qwen1.5-7B}
\end{subfigure}
% \vspace{-1pt}
\caption{Ablation for Pos2Distill-R\textsuperscript{1}. Each point shows mean accuracy, averaged across gold document positions {1,5,10,15,20}. The  $x$-axis represents the total size of training data, which increases with the number of questions, while $\mathcal{\mathcal{K}}$ is fixed.  "KL" refers to word-level KD; adaptive refers to our approach.}
\vspace{-.3cm}
\label{fig: ablations for pos2-1}
\end{figure}

\begin{table}[h!]
    \centering
    \small
    \renewcommand\arraystretch{1}
    \setlength{\tabcolsep}{4pt}
    \resizebox{0.9\columnwidth}{!}{
    \begin{tabular}{@{}lcccccc@{}}
    \toprule
Method&1&5 &10 &15 &20&AVG.\\
\midrule
Base &73.6& 57.3 &56.9& 57.3 &60.9&61.2\\
SeqKD &61.7& 58.7&58.9& 59.9 &61.9& 60.2\\
KL& 61.7&64.3&65.6&66.4&67.6&65.1\\
% -&\checkmark&$\times$& &64.3&65.6&66.4&67.6&\\
KL+Align &64.5&66.5&67.8&67.3&67.4&66.7\\
KL+Align+Anc.&69.9& 67.3& 68.1 &69.1 &67.5 &68.4 \\
    \bottomrule
    \end{tabular}
    }
    \caption{Ablation study of two core modules for Qwen1.5-7B-Chat: the adaptive strategy targeting trivial positions and the anchoring strategy for advantageous positions. The hyperparameter $\lambda$ is set to 1. }
    \label{tab: component ablation}
    \vspace{-4pt}
\end{table}
\subsubsection{Ablation for Pos2Distill-R\textsuperscript{1}} 

% To demonstrate the effectiveness of our design, we respectively ablate two core components namely Position-Aware Alignment and Anchoring of Advantageous Positions. 
% As shown in Tab.~\ref{tab: component ablation}, 我们的Position-Aware Alignment的通过自适应策略在达到更加平衡的性能的同时，进一步优化了学习过程，使模型平均得分增至66.7。最终，结合自适应策略和针对优势位置的锚定策略，模型性能达到最佳，平均得分为68.3。这表明，自适应策略通过优化对次要信息的处理提升了效率，而锚定策略则通过强化关键信息的学习增强了模型的鲁棒性。两种策略的协同作用，显著提升了Qwen1.5-7B-Chat的整体表现，凸显了针对不同类型数据采用差异化学习机制的重要性。

% 为了证明我们设计的有效性，我们针对各个组件进行了详细的消融实验，并得到以下几个重点结论：
To verify the effectiveness of our design, we conducted comprehensive ablation studies in Tab.~\ref{tab: component ablation}. Our findings yield several important insights:

\noindent\textbf{KL is effective for token-shifting correction.} When trained on identical data compositions, the inferior performance of SeqKD highlights the shortcomings of hard-label supervision in scenarios involving token-shifting. In contrast, KL offer a superior mechanism for correcting such shifts compared to the rigid guidance This property makes KL loss particularly well-suited for token-recovery.

\noindent\textbf{Position-aware alignment ensures balance and better learning.} 
Integrating our Position-Aware Alignment strategy with KL divergence leads to significantly more balanced and robust performance, increasing the average score to 66.7.

\noindent\textbf{Anchoring strategy reinforces robustness via key position focus.}
Incorporating anchoring not only addresses attention dilution at sink positions, but also yields performance gains across other positions. With an average score of 68.4, the strategy is particularly effective at position 1 while maintaining strong performance at trivial positions.

% 当将Position-Aware Alignment策略与KL结合时，略在达到更加平衡的性能的同时，进一步优化了学习过程，使模型平均得分增至66.7

% \noindent\textbf{The Efficacy of Adaptive Strategy.} The adaptive strategy always outperforms the word-level KD across various datasets and LLMs, which illustrates adapt to complex patterns, enabling more effective learning of simpler ones later and ensuring balanced learning across varying example complexities.

% The underlying hypothesis is that prioritizing the learning of difficult examples early allows models to adapt to complex patterns, thereby facilitating more effective learning of simpler examples later.  This approach supports balanced learning across different levels of example complexity.
\begin{table*}[t!]
\centering
\small
\renewcommand\arraystretch{0.7}
\setlength{\tabcolsep}{1.8pt}
\resizebox{\linewidth}{!}{
\begin{tabular}{lcccccccccc}
\toprule
 &
\multicolumn{4}{c}{\textbf{MuSiQue}} &
\multicolumn{3}{c}{\textbf{2WikiMultiHopQA}} &
\multicolumn{3}{c}{\textbf{HotpotQA}} \\
\cmidrule(r){2-5} \cmidrule(r){6-8} \cmidrule(r){9-11}
\multicolumn{1}{c}{\textsc{Llama-3.1-8B}}& Overall & 2-Hop & 3-Hop & 4-Hop & Overall & 2-Hop & 4-Hop & Overall & Bridge & Comparison\\
% &\multicolumn{10}{c}{\textit{Zero-Shot Prompting}} \\
\midrule
 +CoT & 11.8&11.9&11.9&7.7&27.4&23.4&40.8&19.4&19.3&23.7 \\
 +CoC & 13.2 & 13.3 & 14.7 & 11.1 &33.8&38.0&50.5&23.2&21.8&29.5 \\
% + \textsc{Fact-and-reflection} & 27.8 & 33.1 & 23.0 & 16.8 & 54.3 &49.5  & 72.6 & 51.8 & 49.6 & 60.3 \\
% \midrule
% &\multicolumn{10}{c}{\textit{Superivised Fine-tuning/Preference Optimization}} \\
% \midrule
+\textsc{Sealong} &25.5&31.6&24.5&15.3&52.6&48.6&66.3&49.6&47.7&56.7\\
+\textsc{LongFaith}-SFT (CoT) &39.9&43.5&37.5&29.1&55.0&49.6&72.9&49.8&50.5&46.7 \\
+\textsc{LongFaith}-SFT (CoC) &39.6&44.4&38.3&29.9&56.6&50.9&75.1&50.9&51.3&50.7 \\
+\textsc{LongFaith}-DPO&34.2&38.8&32.9&25.8&51.2&48.8&59.8&49.7&49.8&42.2 \\
\rowcolor[rgb]{ .867, .922, .969}\textsc{+Pos2Distill} &\textbf{42.8\textsubscript{+2.9}} &\textbf{47.4\textsubscript{+3.0}} &\textbf{38.4\textsubscript{+0.1}} &\textbf{31.6\textsubscript{+1.7}}
&\textbf{61.8\textsubscript{+5.2}}
&\textbf{57.0\textsubscript{+6.1}}
&\textbf{78.0\textsubscript{+2.9}}
&\textbf{58.3\textsubscript{+7.4}}
&\textbf{56.5\textsubscript{+5.2}}
&\textbf{65.8\textsubscript{+9.1}}\\
% + \textsc{Pos2Distill}-DPO & \\
% + \textsc{Pos2Distill}* & 43.2&47.3&36.7&35.2&56.3&53.1&73.2&53.6&53.2&62.4\\
\midrule
\multicolumn{1}{c}{\textsc{Qwen2.5-7B}}& Overall & 2-Hop & 3-Hop & 4-Hop & Overall & 2-Hop & 4-Hop & Overall & Bridge & Comparison\\
\midrule
 +CoT  &28.7&30.4&29.1&25.2&49.5&41.9&76.6&52.1&49.1&62.5\\
 +CoC  &25.9&26.3&28.6&23.7&45.9&38.0&74.3&47.3&43.7&60.6\\
% + \textsc{Fact-and-reflection} & 14.4&14.5&15.7&13.8 &22.8\\
% \midrule
% &\multicolumn{10}{c}{\textit{Superivised Fine-tuning/Preference Optimization}} \\
% \midrule
+\textsc{Sealong}& 33.0&34.5&31.1&28.6 &47.6&42.3&68.0&48.6&48.6&42.1\\
% + \textsc{Sealong}& 15.0&14.5&17.1&12.6 &18.0&15.8&29.0&24.6&28.6&8.0\\
+\textsc{LongFaith}-SFT(CoT) &43.3&45.9&41.1&37.0&51.1&46.2&71.2&53.6&55.7&46.0\\
+\textsc{LongFaith}-SFT(CoC) &42.5&43.3&41.1&38.0&51.1&46.6&68.9&53.6&55.4&47.5\\
+\textsc{LongFaith}-DPO&38.4&39.9&35.7&31.9&59.6&52.5&85.0&56.5&54.2&67.4  \\
% + \textsc{Pos2Distill}-DPO  \\
\rowcolor[rgb]{ .867, .922, .969}\textsc{+Pos2Distill}&\textbf{46.2\textsubscript{+2.9}}&\textbf{47.6\textsubscript{+1.7}}&\textbf{43.5\textsubscript{+2.4}}&\textbf{39.2\textsubscript{+1.2}}&\textbf{63.4\textsubscript{+3.8}}&\textbf{56.7\textsubscript{+4.2}}&76.5&\textbf{58.7\textsubscript{+5.5}}&\textbf{61.2\textsubscript{+7.0}}&67.2  \\
\bottomrule
% \midrule
\end{tabular}}
\caption{Main experiment results on long-context multi-hop reasoning datasets using the \textbf{EM} metric. The training set has 2K samples. The dataset of PosDistill consists 500 questions, setting $\mathcal{K}$ as 4. CoC refers an effective prompting strategy named \textit{Chain-of-Citation} \cite{li2024making}.}
\label{tab: pos2 performance improve}
\vspace{-4pt}
\end{table*}

\subsubsection{Analysis Results}\label{data efficiency}
We investigate the property of Pos2Distill-R\textsuperscript{1}, presenting comprehensive results in Fig.~\ref{fig: ablations for pos2-1}.
\noindent\textbf{High Data Efficiency.} As shown in Fig.~\ref{fig: ablations for pos2-1}, our positional awareness metrics achieve superior performance with minimal training data (e.g., Mistral-7B achieves 70.2\% accuracy with just 250 examples) and consistently outperform other strategies as the dataset grows. This highlights the data efficiency of our approach. As discussed before, this efficiency stems from the distribution similarity in responses, enabling rapid adaptation to in-domain data, avoiding reliance on entirely out-of-domain samples.
\begin{figure}[h!]
    \centering
    \begin{subfigure}{0.49\linewidth}
        \includegraphics[width=\linewidth]{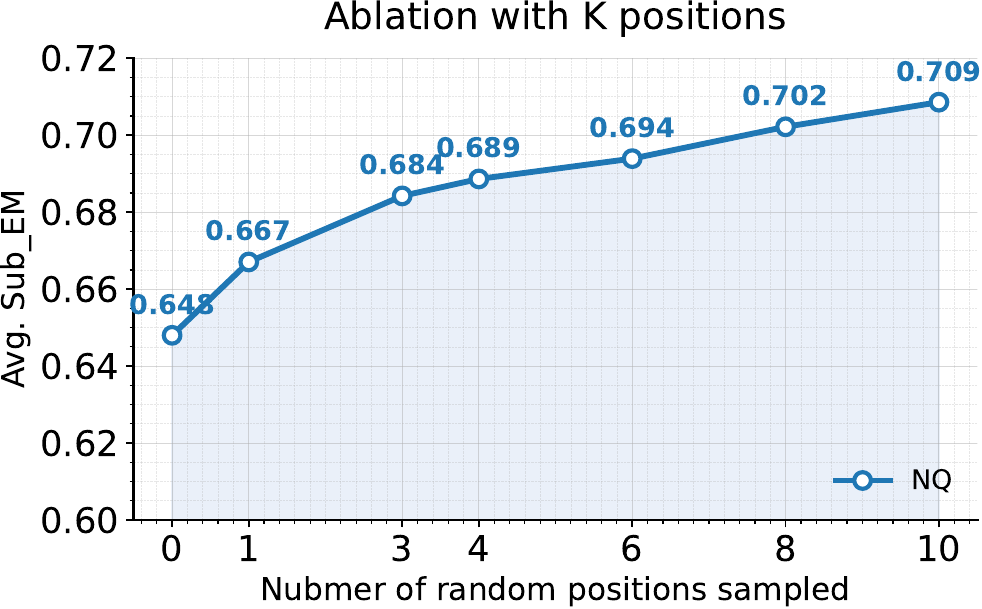}
    \end{subfigure}
    \hfill
    \begin{subfigure}{0.49\linewidth}
        \includegraphics[width=\linewidth]{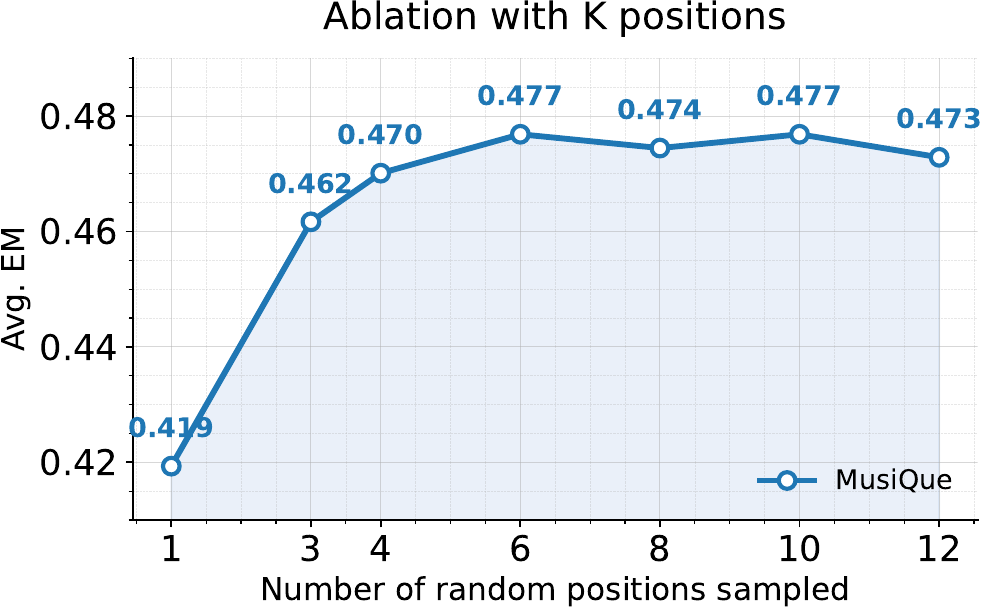}
    \end{subfigure}
    \caption{Ablation for the  number of sampled positions $\mathcal{K}$ for Pos2Distill-R\textsuperscript{1} (left) and Pos2Distill-R\textsuperscript{2} (right).}
    \label{fig: ablations for pos2-2}
\end{figure}

\noindent\textbf{The Impact of $\mathcal{K}$.}
We conduct studies on $\mathcal{K}$ by varying its value from 1 to 10 while keeping other settings fixed. Increasing $\mathcal{K}$ from 0 to 6 improves LLM performance by 4.6\% improvement in LLM performance, but further increase to 10 shows marginal gains with performance saturation. To balance effectiveness and computational efficiency, we select $\mathcal{K}=4\sim6$ as the optimal configuration.

\paragraph{Mechanistic Insights.}

\begin{figure}[h]
    \centering
    \begin{subfigure}{0.49\linewidth}
        \includegraphics[width=\linewidth]{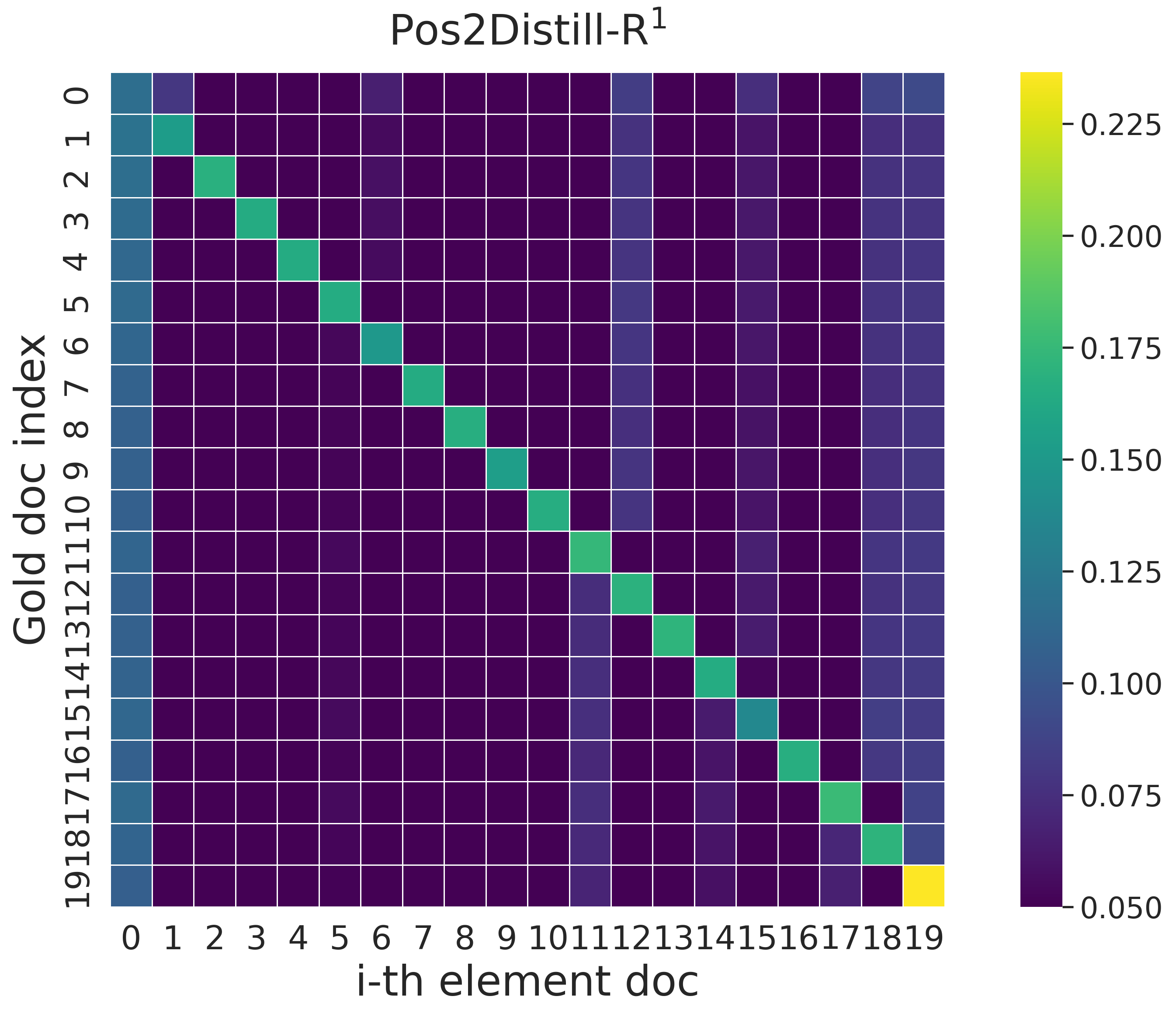}
    \end{subfigure}
    \hfill
    \begin{subfigure}{0.49\linewidth}
        \includegraphics[width=\linewidth]{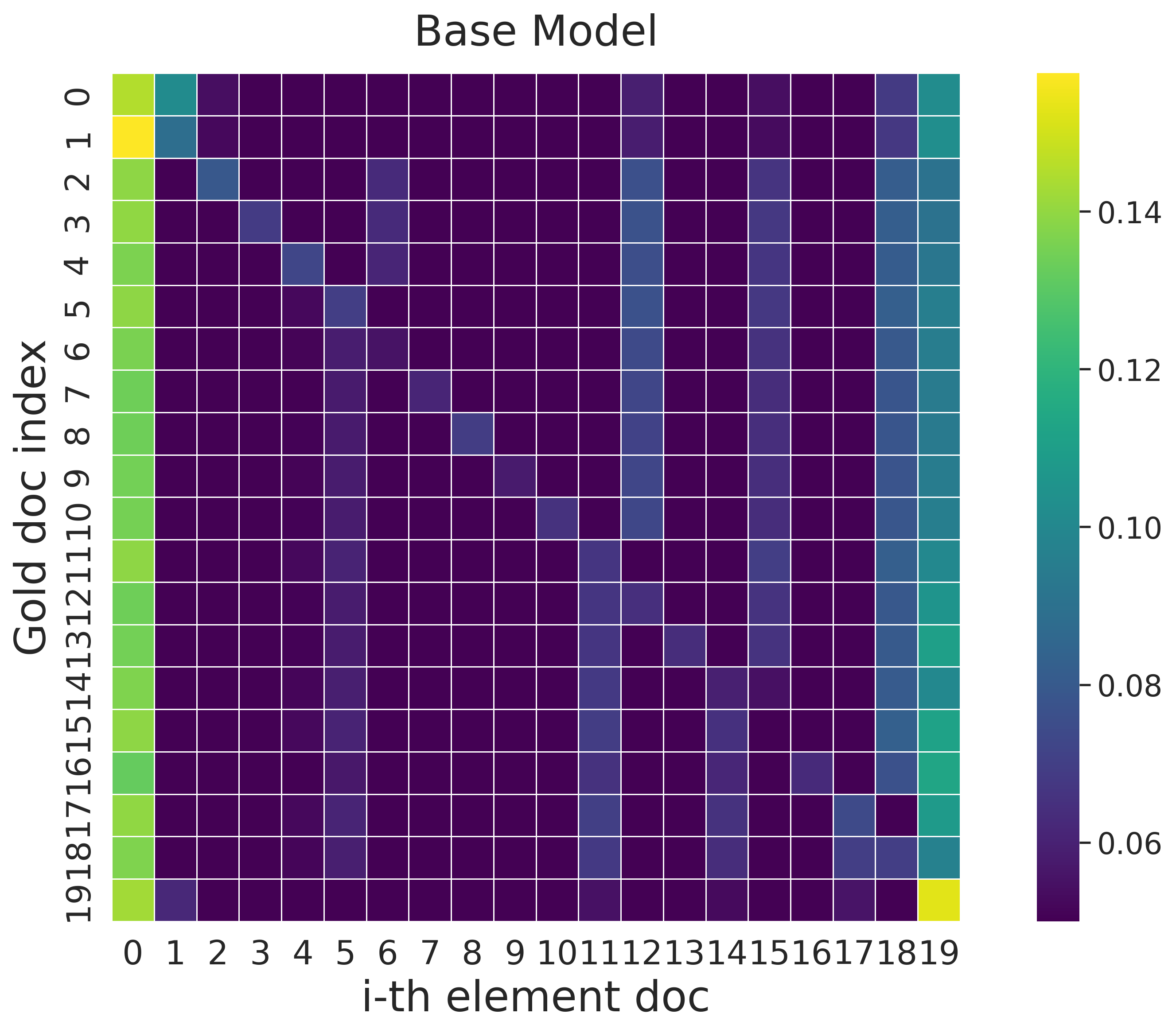}
    \end{subfigure}
    \caption{Attention distribution of each doc across total 20 docs, as the position of $d_{gold}$ varies from 1 to 20.}
    \label{fig: attention vis}
    \vspace{-2pt}
\end{figure}

Since  PB emerges from the architectures and parameters of LLMs, we seek to uncover the internal model dynamics following Pos2Distill-R\textsuperscript{1} and provide an interpretable explanation. We record the attention distribution over 20 documents as $d_{\text{gold}}$ moves from 1 to 20 in Fig.~\ref{fig: attention vis}.
Pos2Distill-R\textsuperscript{1} strengthens contextual fidelity by dynamically shifting the focus of attention to consistently align with the relevant document, thereby facilitating more accurate retrieval.

\begin{table*}[ht!]
\centering
\small
\renewcommand\arraystretch{1}
\setlength{\tabcolsep}{1.8pt}
\resizebox{1\linewidth}{!}{
    \begin{tabular}{lccccc|ccccc|ccccc|c}
    \toprule
    &\multicolumn{5}{c|}{\textbf{Connected}}&\multicolumn{5}{c|}{\textbf{Disconnected}}&\multicolumn{5}{c|}{\textbf{Reversed}}\\
    \cmidrule(r){2-17}
    % \cmidrule(r){2-6} \cmidrule(r){7-11} \cmidrule(r){12-16}
    \textsc{Positions}&[0,1]&[5,6]&[12,13]&[17,18]&\textbf{Avg.$\uparrow$ }&[0,8]&[5,13]&[6,14]&[8,16]&\textbf{Avg.$\uparrow$ }&[8,0]&[13,5]&[14,6]&[16,8]&\textbf{Avg.}&\textbf{\textsc{Gap.}	$\downarrow$  } \\
    \midrule
    % \textsc{Qwen2.5-3B}&21.8&18.8&19.2&20.5&\textbf{20.1}&19.8&15.3&16.0&16.5&\textbf{16.9}&18.9&16.4&16.6&13.9 &\textbf{16.5}&7.9\\
    % \rowcolor[rgb]{ .867, .922, .969}\textsc{+pos2distill}&  \\
    \textsc{Qwen2.5-7B}& 37.4&33.2&32.9&38.4&\textbf{35.5}&34.8&26.8&28.2&27.7&\textbf{29.4}&33.3&30.2&29.6&29.8&\textbf{30.7}&11.6 \\
    \textsc{+longfaith-sft}& 51.0&46.6&46.7&50.1&\textbf{48.6}&47.1&44.3&44.2&43.8&\textbf{44.9}&45.4&42.6&41.7&43.0&\textbf{43.2}&9.3\textsubscript{-2.3}\\
    \textsc{+longfaith-dpo}&39.0&31.1&35.5&40.5&\textbf{36.5}&37.3&32.6&31.5&30.4&\textbf{33.0}&39.0&31.1&35.5&40.5&\textbf{36.5}&10.1\textsubscript{-1.5} \\
    \rowcolor[rgb]{ .867, .922, .969}\textsc{+pos2distill}&50.6&49.8&49.1&51.3&\textbf{50.3}&48.5&48.1&47.3&46.7&\textbf{47.4}&47.7&46.5&47.1&47.3&\textbf{47.2}&4.8\textsubscript{-6.8}  \\
    % \textsc{Qwen2.5-14B}&46.8&43.8&43.4&46.6&\textbf{45.2}&46.0&40.0&40.0&41.4&\textbf{41.9}&47.5&41.4&40.9&40.2&\textbf{42.5}&7.5 \\
    
    % \rowcolor[rgb]{ .867, .922, .969}\textsc{+pos2distill}&  \\
    % \textsc{Llama3.2-3B}& 23.1&21.1&23.4&24.9&\textbf{23.1}&23.1&21.5&21.8&20.5&\textbf{21.7}&21.0&19.5&19.7&19.2&\textbf{19.8}&5.7\\
    % \rowcolor[rgb]{ .867, .922, .969}\textsc{+pos2distill}&  \\
    \textsc{Llama3.1-8B}& 14.3&13.4&15.4&16.0&\textbf{14.8}&15.8&15.5&13.7&14.5&\textbf{14.9}&14.3&11.6&12.9&12.3&\textbf{12.8}&4.4 \\
    \textsc{+longfaith-sft}&45.4&44.4&44.9&47.4&\textbf{45.5}&44.3&45.2&43.5&42.7&\textbf{43.9}&42.9&42.7&42.9&41.5&\textbf{42.5}&5.9\textsubscript{+1.5}  \\
    \textsc{+longfaith-dpo}& 37.4&38.4&40.7&40.1&\textbf{39.2}&37.7&37.4&36.6&36.2&\textbf{37.0}&34.8&34.2&36.4&36.0&\textbf{35.4}&6.5\textsubscript{+2.1} \\
    \rowcolor[rgb]{ .867, .922, .969}\textsc{+pos2distill}& 47.2&48.4&47.8&49.4&\textbf{48.2}&47.5&49.0&47.2&46.6&\textbf{47.6}&45.6&46.0&46.1&47.0&\textbf{46.2}&3.4\textsubscript{-1.0} \\
    \bottomrule

    \end{tabular}}
    \caption{PB assessment for Pos2Distill-R\textsuperscript{2} on two-hop data from MusiQue.}
    \vspace{-.3cm}
    \label{tab:pos2 pb}
\end{table*}
\subsection{Main Results for Pos2Distill-R\textsuperscript{2}}
\paragraph{Pos2Distill-R\textsuperscript{2} strengthens in-context reasoning.}
Pos2Distill-R\textsuperscript{2} surpasses existing self-training approaches in both in-domain performance and out-of-domain generalization. 
As detailed in Tab.~\ref{tab: pos2 performance improve}, when trained on the MusiQue dataset, Pos2Distill-R\textsuperscript{2} achieves an Exact Match (EM) score of 42.8, outperforming all leading baselines. Furthermore, our method exhibits robust cross-domain generalization; for instance, on the HotpotQA dataset, it attains an EM score of 58.3, compared to 50.9 from the strongest baseline.
% These results highlight the efficacy of addressing PB for the CoT paradigm, which proves to be more effective than just teaching LLMs learning to reason over context one instance by instance.
Our findings suggest that \textit{training LLMs to reason across diverse, scattered gold positions potentially enhances their long-context reasoning more effectively than conventional instance-by-instance training.} 
This insight offers a new perspective for improving reasoning capabilities in complex, long-context tasks.

To assess PB within the reasoning paradigm, we evaluate performance on two-hop data from MusiQue, considering three relative position configurations for the two gold documents: (i) \emph{connected} (hops are adjacent), (ii) \emph{disconnected} (distracting content separates the hops), and (iii) \emph{reversed} (hops are logically inverted and disconnected).
% When multiple gold documents are present, PB depends not only on their positions within the context window, but also on the inter-document distance and their relative ordering.
Illustratively, the accuracy of Qwen2.5-7B  peaks at 38.4\% in the connected mode (hops  at positions 17 and 18) but declines to 26.8\% in the disconnected mode (hops at positions 5 and 13), exhibiting a obvious performance gap 11.6\%. As detailed in Tab.~\ref{tab:pos2 pb}, our method not only effectively mitigates the performance gaps to 4.1\% across these three modes but also enhances overall reasoning performance irrespective of hop configuration. Conversely, conventional self-training methods, which typically rely on instance-by-instance learning, struggle to eliminate these inter-mode disparities and may even exacerbate this trend, which underscores a fundamental limitation of their training paradigm in handling such PB.

\subsubsection{Analysis for Pos2Distill-R\textsuperscript{2}}

\textbf{The Impact of $\mathcal{K}$ on \textbf{Pos2Distill-R\textsuperscript{2}}.} 
In Fig.~\ref{fig: ablations for pos2-2} (right),
a similar trend to Pos2Distill-R\textsuperscript{1} is observed: initially increasing $\mathcal{K}$ leads to a notable improvement  in EM scores, but subsequently,  but further increases result in diminishing returns, with performance eventually reaching a saturation point.
Therefore, an optimal range for $\mathcal{K}$ exists, beyond which further increases yield marginal benefits.
\subsection{In-depth Exploration}
\begin{table}[h!]
    \centering
    \small
    \tabcolsep 1.1pt
    \resizebox{0.95\columnwidth}{!}{
    \begin{tabular}{lcccc}
    \toprule
    &\multicolumn{2}{c}{\textbf{Retrival}}&\multicolumn{2}{c}{\textbf{Reasoning}} \\
    \cmidrule(r){2-3} \cmidrule(r){4-5}
\textbf{Task}&\textbf{NQ}&\textbf{WebQ}&\textbf{MusiQue}&\textbf{HotpotQA}\\
    \midrule
    % &\multicolumn{4}{c}{\textbf{\textsc{Pos2Distill-R\textsuperscript{1}}}} \\
    % \cmidrule(r){2-5}
    \textsc{Base}&62.5&63.9&41.8&66.7\\
    Pos2Distill-R\textsuperscript{1}&74.3\textsubscript{\textbf{+11.8}}&68.1\textsubscript{\textbf{+4.2}}&45.1\textsubscript{+3.3}&69.2\textsubscript{+2.5}\\
    % &\multicolumn{4}{c}{\textbf{\textsc{Pos2Distill-R\textsuperscript{2}}}} \\
    % \cmidrule(r){2-5}
    % \textsc{CoT}&56.4&56.7\\  
    Pos2Distill-R\textsuperscript{2}&64.1\textsubscript{+1.6}&66.7\textsubscript{+2.8}&48.9\textsubscript{\textbf{+7.1}}&72.3\textsubscript{\textbf{+5.6}}\\
    \bottomrule
    \end{tabular}}
    \caption{Performance of Qwen1.5-7B fine-tuned with Pos2Distill-R\textsuperscript{1} versus  with Pos2Distill-R\textsuperscript{2} on retrieval and reasoning tasks, utilizing the same size of data.}
    \label{tab: comparision between rs}
    \vspace{-.1in}
\end{table}

\begin{table*}[!h]
\centering
\small
\setlength{\tabcolsep}{2pt} % 调整列之间的间隔
\renewcommand{\arraystretch}{0.85} % 减小表格行间距
\vspace{-.1in} % 压缩表格与上文之间的空间
\begin{tabular}{lcccccccc|cccccccc}
\toprule
\textbf{Model} & \multicolumn{8}{c}{\textbf{NQ}} & \multicolumn{8}{c}{\textbf{Webq}} \\
\cmidrule{2-8} \cmidrule{11-17}
 & 0    & 5    & 10   & 15   & 20   & 25   & 30   &  \textbf{Gap.} $\downarrow$   & 0  & 5    & 10   & 15   & 20   & 25   & 30   &  \textbf{Gap.} $\downarrow$   \\
\midrule
Qwen2.5-14B    & 72.8 & 68.9 & 64.7 & 67.5 & 66.5 & 64.7 & 69.7 & 8.1   & 64.2 & 59.6 & 59.2 & 60.8 & 59.2 & 57.0 & 59.1 & 7.2   \\
\rowcolor[rgb]{ .867, .922, .969}+Pos2Distill-R\textsuperscript{1} & 71.5 & 68.9 & 70.5 & 71.1 & 71.6 & 72.1 & 72.3 & \textbf{3.4} & 62.0 & 64.4 & 62.2 & 62.4 & 62.6 & 63.8 & 63.2 & \textbf{2.4} \\
Qwen2.5 32B   & 70.3 & 65.1 & 65.5 & 65.9 & 64.3 & 66.1 & 71.3 & 7.0   & 64.2 & 59.6 & 59.2 & 60.8 & 59.2 & 57.0 & 59.0 & 7.2   \\
\rowcolor[rgb]{ .867, .922, .969}+Pos2Distill-R\textsuperscript{1}  & 71.5 & 67.7 & 70.3 & 70.1 & 70.9 & 70.7 & 70.7 & \textbf{3.8} & 63.2 & 63.2 & 61.6 & 60.6 & 61.2 & 62.2 & 62.8 & \textbf{2.6} \\
\bottomrule
\end{tabular}
\caption{Generalization results on model size for Pos2Distill-R\textsuperscript{1}.}
\vspace{-.14in} % 压缩表格与下文之间的空间
\label{tab: generalization_model_size1}
\end{table*}

\begin{table*}[!h]
\centering
\small
\setlength{\tabcolsep}{2pt} % Adjust spacing between columns
\renewcommand{\arraystretch}{1.2} % Adjust row height to make it look nice
\begin{tabular}{lcccc|cccc|cccc|cc}
\toprule
\textbf{Musique} & \multicolumn{4}{c}{\textbf{Connected}} & \multicolumn{4}{c}{\textbf{Disconnected}} & \multicolumn{4}{c}{\textbf{Reversed}} & \textbf{Gap} $\downarrow$ \\
\cmidrule(lr){2-5} \cmidrule(lr){6-9} \cmidrule(lr){10-13}
 & [0,1] & [5,6] & [12,13] & [17,18] & [0,8] & [5,13] & [6,14] & [8,16] & [8,0] & [13,5] & [14,6] & [16,8] & \\
\midrule
Qwen2.5 14B       & 56.6 & 54.1 & 54.7 & 59.5 & 55.0 & 49.5 & 50.0 & 51.8 & 57.7 & 51.4 & 52.0 & 51.9 & 10.0 \\
\rowcolor[rgb]{ .867, .922, .969}+PosDistill R\textsuperscript{2}  & 60.1 & 58.4 & 60.9 & 63.2 & 59.5 & 56.7 & 57.7 & 58.1 & 59.0 & 58.6 & 56.7 & 56.5 & \textbf{6.7} \\
Qwen2.5 32B       & 61.7 & 59.8 & 59.1 & 63.2 & 59.4 & 54.7 & 54.2 & 54.7 & 60.3 & 56.6 & 55.4 & 57.8 & 9.0 \\
\rowcolor[rgb]{ .867, .922, .969}+PosDistill R\textsuperscript{2}  & 64.2 & 65.2 & 63.1 & 65.7 & 63.4 & 61.8 & 60.8 & 61.0 & 62.8 & 62.9 & 61.8 & 62.4 & \textbf{4.9} \\
\bottomrule
\end{tabular}
\caption{Generalization results on model size for Pos2Distill-R\textsuperscript{2}.}
\label{tab: generalization model size2}
\vspace{-.2in}
\end{table*}
\paragraph{Discussion on Two Systems.} As presented in Tab.~\ref{tab: comparision between rs}, both systems exhibit notable generalization to their mutual tasks.
Specifically, Pos2Distill-R\textsuperscript{1}, primarily optimized for retrieval, demonstrates that its enhanced contextual retrieval capabilities also improve reasoning over long contexts, yielding a 3.3\% increase on the MusiQue task.
Conversely, Pos2Distill-R\textsuperscript{2}, optimized for reasoning, shows that its acquired proficiency in reasoning over long context also bolsters contextual awareness, thereby benefiting retrieval performance.
Despite this cross-task generalization, each system excels in its primary domain: Pos2Distill-R\textsuperscript{2} achieves superior performance on complex long-context reasoning tasks where Pos2Distill-R\textsuperscript{1} lags, and vice versa for retrieval. This suggests distinct underlying dynamics for mitigating PB, potentially influenced by the presence or absence of CoT. Consequently, the development of these two specialized Pos2Distill designs proves both necessary and effective.

\paragraph{Generalization to Larger LLMs}  To further ascertain the robustness and broad applicability of our findings, we extended our investigations to  larger-scale LLMs. The comprehensive evaluations presented in Tab. \ref{tab: generalization_model_size1} and \ref{tab: generalization model size2}, conducted across 30 docs, consistently reveal that even significantly larger models exhibit a pronounced  prevalence of PB. This critical observation underscores the universal nature and persistence of PB across model scales. Notably, our proposed method, Pos2Distill-R\textsuperscript{1} and -R\textsuperscript{2}, proved remarkable effectiveness in mitigating PB within both 14 and 32B. Specifically, for the Qwen2.5-32B, Pos2Distill-R\textsuperscript{1} significantly reduced the performance gap: decreasing it from 7.2\% to 2.6\% in retrieval tasks and from 9\% to 4.9\% in reasoning tasks. These compelling quantitative results  affirm the scalability and efficacy of our method when applied to larger LLMs.
% Our results on models with larger sizes evaluated on 30 docs demonstrate that larger models still exhibit strong PB, for example, Qwen2.5-14B shows a 7.2\% performance gap across positions. \textbf{\textit{This indicates that PB universally exists, even in LLMs with larger size.}}
% Our method (Pos2Distill R1 and R2) effectively alleviates PB in both of them. Specifically, for Qwen2.5-32B, the performance gap decreases from 7.2\% to 2.6\% in retrieval tasks and from 9\% to 4.9\% in reasoning tasks after applying our method. This clearly demonstrates the scalability and efficacy of our framework on larger LLMs, addressing the concern about generalization beyond small models.

\section{Conclusion}
This work introduces a novel paradigm to mitigate PB by leveraging the performance disparity it creates. Specifically, our method distills knowledge from privileged positions to unfavored ones, thereby reducing the disparty induced by PB. PB  manifests as token-shifting in retrieval and as thought-shifting in reasoning. To address distinct facets of PB dynamics, we develop two specialized frameworks: Pos2Distill-R\textsuperscript{1} and Pos2Distill-R\textsuperscript{2}. Extensive experiments validate the efficacy of our approach in reducing PB and robust generalization in both in-context retrieval and reasoning tasks.

\newpage
\section*{Limitations}
While our proposed methods demonstrate substantial improvements in performance and data efficiency, we acknowledge certain limitations exist. Specifically, for Pos2Distill-R\textsuperscript{2}, there is scope for further refinement. The current design, while effective, could benefit from more granular mechanisms to precisely calibrate the mitigation of PB. For instance, future work could explore adaptive strategies that adjust the positional distillation process based on the complexity of the reasoning chain or the specific configuration of supporting documents. Such enhancements might lead to even more nuanced control over positional influences in complex reasoning scenarios.

\section*{Potential Social Impacts}

Enhancing positional robustness in large language models (LLMs) fosters more reliable, fair, and consistent information processing, especially in scenarios requiring long-context retrieval and reasoning. By mitigating PB, our approach encourages equitable model behavior and reduces spurious disparities in output quality that could disadvantage critical content occurring in less prominent positions. These improvements are especially significant for real-world applications where fair and accurate comprehension of lengthy documents is essential, such as in education, law, healthcare, and scientific research. In these settings, a focus on understanding and reasoning supports the development of more inclusive and trustworthy AI systems, enabling better information access and more dependable, model-assisted decision making. Ultimately, our work advances large language models as robust and ethical tools for the benefit of society.

\section*{Acknowledgement} 

This work was supported in part by the Strategic Priority Research Program of Chinese Academy of Sciences under Grant XDA0480301.

\bibliography{custom}

\clearpage
\newpage
\appendix

\newpage
\section{Related Work} \label{appendix related work}

\paragraph{On the Emergence of PB.}
This section elucidates the fundamental mechanisms governing position bias. 
Position bias describes the tendency to prioritize information differentially based on its position within an input. A notable example of this effect is the "lost-in-the-middle" phenomenon, characterized by a pronounced performance gap depending on the placement of crucial information. Specifically, models tend to achieve higher accuracy when key information is positioned at the sequence boundaries but experience significant performance degradation when it is located in the middle. This pattern emerges despite strong performance at both ends of the sequence, driven by a combination of primary bias, which favors early content, and recency bias, which enhances reliance on recent information.

\textbf{Primary Bias.}
The enhanced utilization of initial context can be attributed to a universal phenomenon known as ``attention sinks''. This concept refers to the inherent tendency of models to allocate substantial attention to initial tokens, regardless of their semantic relevance. Consequently, LLMs often give disproportionate emphasis to the first few tokens in a sequence. The emergence of attention sinks can be traced back to the intricate interplay of pre-training dynamics, encompassing factors such as optimization processes, data distribution, and the model's loss function \cite{gu2025when}. 
\[
\begin{aligned}
\textbf{RoPE:} \quad &\mathbf{q}_m = \mathcal{R}(\mathbf{q}, m), \quad \mathbf{k}_n = \mathcal{R}(\mathbf{k}, n) \\
&\mathbf{q}_m \mathbf{k}_n^{\mathrm{T}} = \mathcal{G}(\mathbf{q}, \mathbf{k}, m - n) \\
\textbf{Attention:} \quad 
&\mathbf{a}_{m,n} = \text{Softmax} \left( \frac{\mathbf{q}_m \mathbf{k}_n^{\mathrm{T}} + \mathbf{mask}}{\sqrt{d}} \right) 
\end{aligned}
\]

\textbf{Recency Bias.}
 The preferential attention allocation to terminal positions arises from two key architectural components: \textbf{Causal Mask} and \textbf{Relative Position Encoding}. First, the causal mask enforces a unidirectional flow of information ($m > n$), implicitly embedding positional information and producing context-dependent token embeddings that vary with sequential permutations. Simultaneously, Rotary Position Embeddings (RoPE) encode relative positional relationships ($m-n$ in equations) within attention calculations, inherently biasing the model toward recent tokens. While the intricate interplay between these components warrants further investigation, this preliminary analysis highlights the emergence of asymmetric attention patterns, providing insights that inform both the design and broader understanding of the framework.
 
 The advent of the long-context era for Large Language Models (LLMs) has been notably advanced by progress in two pivotal directions: (1) efficient attention mechanisms such as FlashAttention \cite{dao2022flashattention, ainslie2023gqa}, which drastically reduce the computational overhead of processing extended sequences, and (2) length extrapolation techniques \cite{he2024two, chi-etal-2023-dissecting, press2021train}, which enable LLMs to generalize beyond their training context length \cite{su2024roformer, raffel2020exploring}. Collectively, these breakthroughs empower LLMs to perform complex question answering over much larger context windows \cite{agarwal2024manyshot}, representing a significant step toward more capable and flexible natural language systems~\cite{yang2025unearthing}. 
 
\paragraph{Mechanistic Approaches.} For instance, to alleviate intrinsic long-range decay~\cite{zhao2025gdllm}, \citeauthor{zhang2024found} propose \textit{Ms-PoE}, which assigns distinct rescaling factors to each attention head, compressing relative distance $m-n$ by a factor of $1/n$. The work \textit{Attention Buckets} \cite{chen-etal-2024-fortify} exploits the approximate periodicity observed in the attention waveform at distal positions, shifting key information away from the waveform’s trough regions. Furthermore, \textit{MoICE} \cite{lin2024mixture} designs a router within each attention head to dynamically select among multiple RoPE angles, effectively avoiding trough zones during generation. \citeauthor{yu2024mitigate} reveals that PB is reflected in positional hidden states, and mitigates this by scaling certain dimensions of these representations. However, these component-level modifications without continual training present two limitations: (1) obtaining optimal hyperparameters requires multiple forward passes~\cite{chen-etal-2024-fortify}; and (2) such interventions disrupts the parallelism of multi-head attention \cite{zhang2024found}.

\paragraph{Training Approach.} Therefore, FILM~\cite{an2024make} systematically synthesizes long-context QA datasets to promote fine-grained information awareness, encompassing a range of in-context retrieval and reasoning tasks over varying context lengths. Although such training allows LLMs to better utilize extended contexts, it incurs considerable computational overhead. \citet{zhang2024positionawareparameterefficientfinetuning} introduce a position-aware adapter module to remove intrinsic preference bias. However, this approach is limited to single-segment retrieval and does not generalize well to more complex scenarios involving integration and reasoning over multiple segments.
\section{Preliminary Results}
\label{appendix: Additional Preliminary Results}
\subsection{Token Shifting Phenomenon} \label{appendix: complete responses}
\textbf{Gold index 1}: 
\textcolor{red}{The first orca to be put in captivity was} Wanda, a North Eastern Pacific killer whale, which was captured in \textcolor{blue}{November 1961} by a collecting crew from Marineland of the Pacific in Los Angeles. \\
\textbf{Gold index 2}: \textcolor{red}{The first orca to be put in captivity was} Wanda, a North Eastern Pacific killer whale, who was captured in \textcolor{blue}{November} 1961 by a collecting crew from Marineland of the Pacific in Los Angeles. \\
\textbf{Gold index 3}: \textcolor{red}{The first orca to be put in captivity was} Wanda, a North Eastern Pacific killer whale, who was captured in \textcolor{blue}{November 1961} by a collecting crew from Marineland of the Pacific in Los Angeles. \\
\textbf{Gold index 4}: \textcolor{red}{The first orca to be put in captivity was} Wanda, a North Eastern Pacific killer whale, who was captured in \textcolor{blue}{November 1961} by a collecting crew from Marineland of the Pacific in Los Angeles. \\
\textbf{Gold index 5}: \textcolor{red}{The first orca to be put in captivity was} Moby Doll, a male orca captured in \textcolor{blue}{November 1961} by a collecting crew from Marineland of the Pacific in Los Angeles. \\
\textbf{Gold index 6}: \textcolor{red}{The first orca to be put in captivity was} Moby Doll, a male orca captured in \textcolor{purple}{1964} by Ted Griffin.  Moby Doll was held at the Vancouver Aquarium for three months before dying.\\
\textbf{Gold index 7}: \textcolor{red}{The first orca to be put in captivity was} Moby Doll, a male orca captured in \textcolor{purple}{1964} by Ted Griffin. Moby Doll was held at the Vancouver Aquarium for three months before dying.\\
\textbf{Gold index 8}: \textcolor{red}{The first orca to be put in captivity was} Moby Doll, a male orca captured in \textcolor{purple}{1964} by Ted Griffin.  Moby Doll was held at the Vancouver Aquarium for three months before dying.\\
\textbf{Gold index 9}: \textcolor{red}{The first orca to be put in captivity was} Moby Doll, a male orca captured in \textcolor{purple}{1964} by a collecting crew from Marineland of the Pacific in Los Angeles.\\
\textbf{Gold index 10}: \textcolor{red}{The first orca to be put in captivity was} Moby Doll, a male orca captured in \textcolor{purple}{1964} by a collecting crew from Marineland of the Pacific in Los Angeles. He was placed in a tank at the aquarium.\\
\textbf{Gold index 11}: \textcolor{red}{The first orca to be put in captivity was} Moby Doll, a male orca captured in \textcolor{purple}{1964} by a collecting crew from Marineland of the Pacific in Los Angeles. \\
\textbf{Gold index 12}: \textcolor{red}{The first orca to be put in captivity was} Moby Doll, a male orca captured in \textcolor{purple}{1964} by a collecting crew from Marineland of the Pacific in Los Angeles. \\
\textbf{Gold index 13}: \textcolor{red}{The first orca to be put in captivity was} Moby Doll, a male orca captured in \textcolor{purple}{1964} and displayed at the Vancouver Aquarium for three months before dying.  \\
\textbf{Gold index 14}: \textcolor{red}{The first orca to be put in captivity was} Moby Doll, a male orca captured in \textcolor{purple}{1964} by a collecting crew from Marineland of the Pacific in Los Angeles.   \\
\textbf{Gold index 15}: \textcolor{red}{The first orca to be put in captivity was} Moby Doll, a male orca captured in \textcolor{purple}{1964} and displayed at the Vancouver Aquarium in British Columbia.    \\
\textbf{Gold index 16}: \textcolor{red}{The first orca to be put in captivity was} Moby Doll, who was captured in \textcolor{purple}{1964} and held at the Vancouver Aquarium for three months before dying.    \\
\textbf{Gold index 17}: \textcolor{red}{The first orca to be put in captivity was} Moby Doll, who was captured in \textcolor{purple}{1964} and displayed at the Vancouver Aquarium for three months.     \\
\textbf{Gold index 18}: \textcolor{red}{The first orca to be put in captivity was} Moby Doll, a male orca captured in \textcolor{purple}{1964} and displayed at the Vancouver Aquarium in British Columbia. \\
\textbf{Gold index 19}: \textcolor{red}{The first orca to be put in captivity was} Moby Doll, who was captured in \textcolor{purple}{1964} and displayed at the Vancouver Aquarium in British Columbia.  \\
\textbf{Gold index 20}: \textcolor{red}{The first orca to be put in captivity was} Moby Doll, who was captured in \textcolor{purple}{1964} and displayed at the Vancouver Aquarium for three months. 

\paragraph{Phenomena Related to PB} To verify the phenomenon mentioned in Sec.~\ref{sec: Behavior of Position Bias}, we provide more proof related to them: (1) \textbf{Token shifting:} We analyze token-level KL divergence between responses generated from trivial and advantageous positions to verify the existence of token-shifting via case study in Tab. \ref{table: phenomenon-related analysis experiments}. Our results on Mistral and Qwen demonstrate that token shifting is indeed a universal phenomenon: KL divergence values at specific token positions is extremely high compared to normal level, indicating that the severe divergence happens at this decoding step, which is named token-shifting in our paper. According to the table, it is obvious token-shifting only take places on very few tokens; (2) \textbf{Token recovery:} Our Pos2Distill-R\textsuperscript{1} framework especially focus on these key tokens (those exhibiting higher KL divergence at token-level) and fix them, in order to recover the optimal decoding trajectory like sink positions again, thus mitigating PB. This core concept parallels recent findings in mathematical reasoning tasks \cite{wang2025beyond}, where tokens with high entropy are labeled as “forking tokens.” Interventions that specifically correct model predictions at these forking tokens have been shown to outperform even full gradient updates; (3) \textbf{Compound PB:} This occurs in reasoning tasks in long-context scenarios and involves the interplay of retrieval and actual thinking processes, leading to very sophisticated compound effects. Therefore, instead of just focusing retrieval or reasoning, we choose to reshape CoT process. Therefore, Pos2Distill-R\textsuperscript{2} can be seen as the form of Reinforced Self-Training \cite{gulcehre2023reinforced}, Self-Taught Reasoner \cite{zelikman2022star} or HS-STAR \cite{xiong2025hs}.

\begin{table*}[htbp]
\centering
\small
\setlength{\tabcolsep}{3pt} % 调整列之间的间距为 3pt
\begin{tabular}{lcccccccccc}
\toprule
\textbf{Response} & \textbf{Token1} & \textbf{Token2} & \textbf{Token3} & \textbf{Token4} & \textbf{Token5} & \textbf{Token6} & \textbf{Token7} & \textbf{Token8} & \textbf{Token9} & \textbf{Token10} \\
\midrule
Mistral+NQ     & 1.6e-03 & 1.4e-02 & 5.7e-06 & 7.7e-05 & 1.87e-03 & \textbf{1.4e-01} & 1.6e-03 & 4.8e-06 & 2.0e-07 & 2.6e-07 \\
Mistral+Webq   & 5.7e-03 & 5.5e-03 & 5.4e-05 & 1.0e-05 & 1.4e-04 & 3.6e-03 & 3.3e-06 & 1.2e-04 & \textbf{1.7e-02} & 4.4e-03 \\
Llama3+NQ      & 5.1e-03 & 6.6e-04 & 6.0e-03 & 2.2e-04 & 7.3e-04 & 1.0e-03 & 2.9e-04 & \textbf{1.0e-02} & 7.3e-06 & 5.3e-05 \\
Llama3+Webq    & 9.6e-03 & 2.6e-04 & 9.1e-03 & 7.7e-08 & 2.9e-04 & 1.6e-04 & 2.1e-06 & \textbf{1.2e-02} & 1.9e-06 & 2.1e-06 \\
\bottomrule
\end{tabular}
\caption{Phenomenon related token-shifting.}
\label{table: phenomenon-related analysis experiments}
\end{table*}

\begin{figure*}[!ht]
    \centering
    \begin{minipage}[t]{0.45\textwidth}
        \centering
        \includegraphics[width=\linewidth]{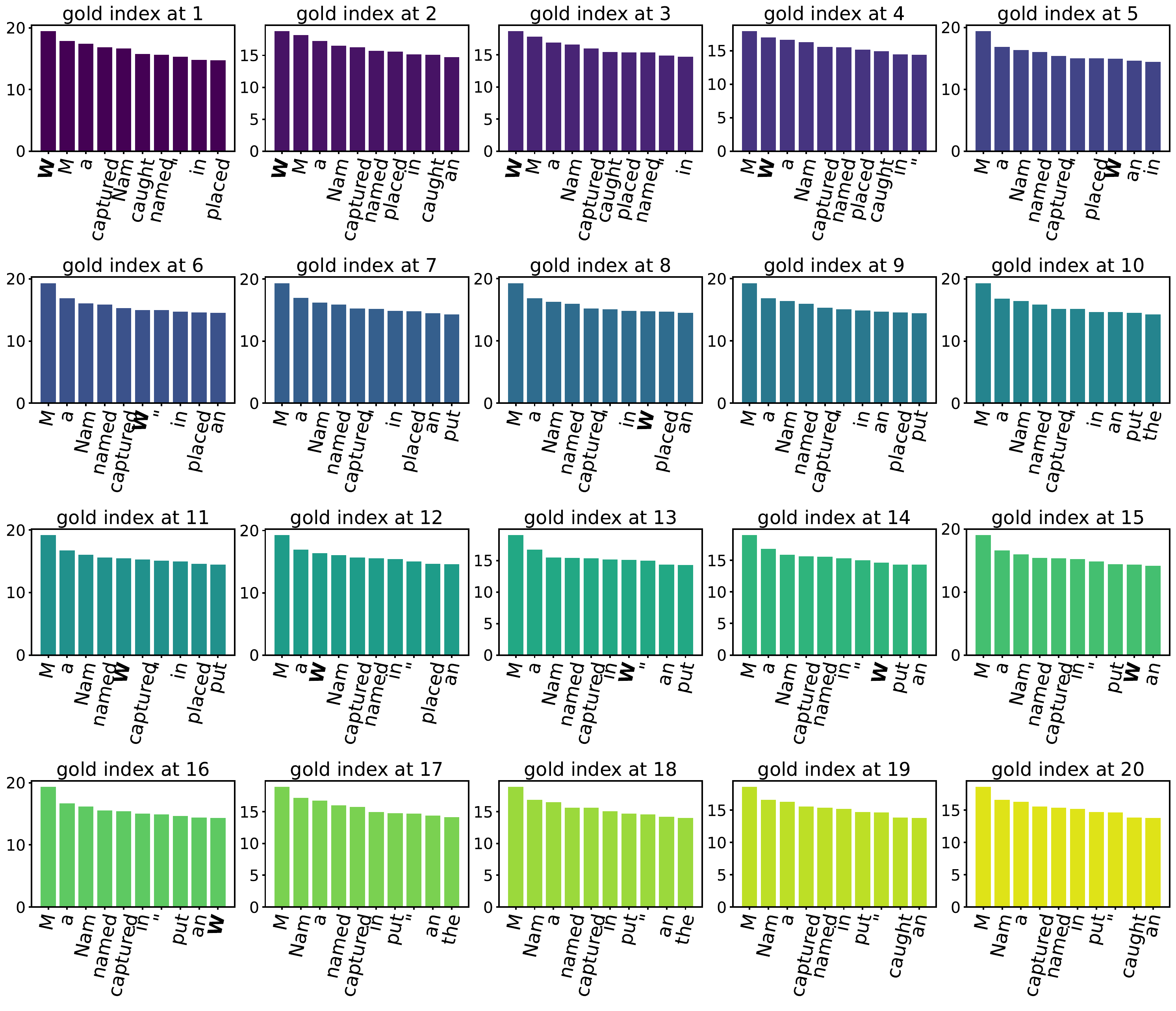}
        \label{fig:Meta-Llama-3-8B-Instruct_logits}
    \end{minipage}%
    \hspace{0.02\textwidth}
    \begin{minipage}[t]{0.45\textwidth}
        \centering
\includegraphics[width=\linewidth]{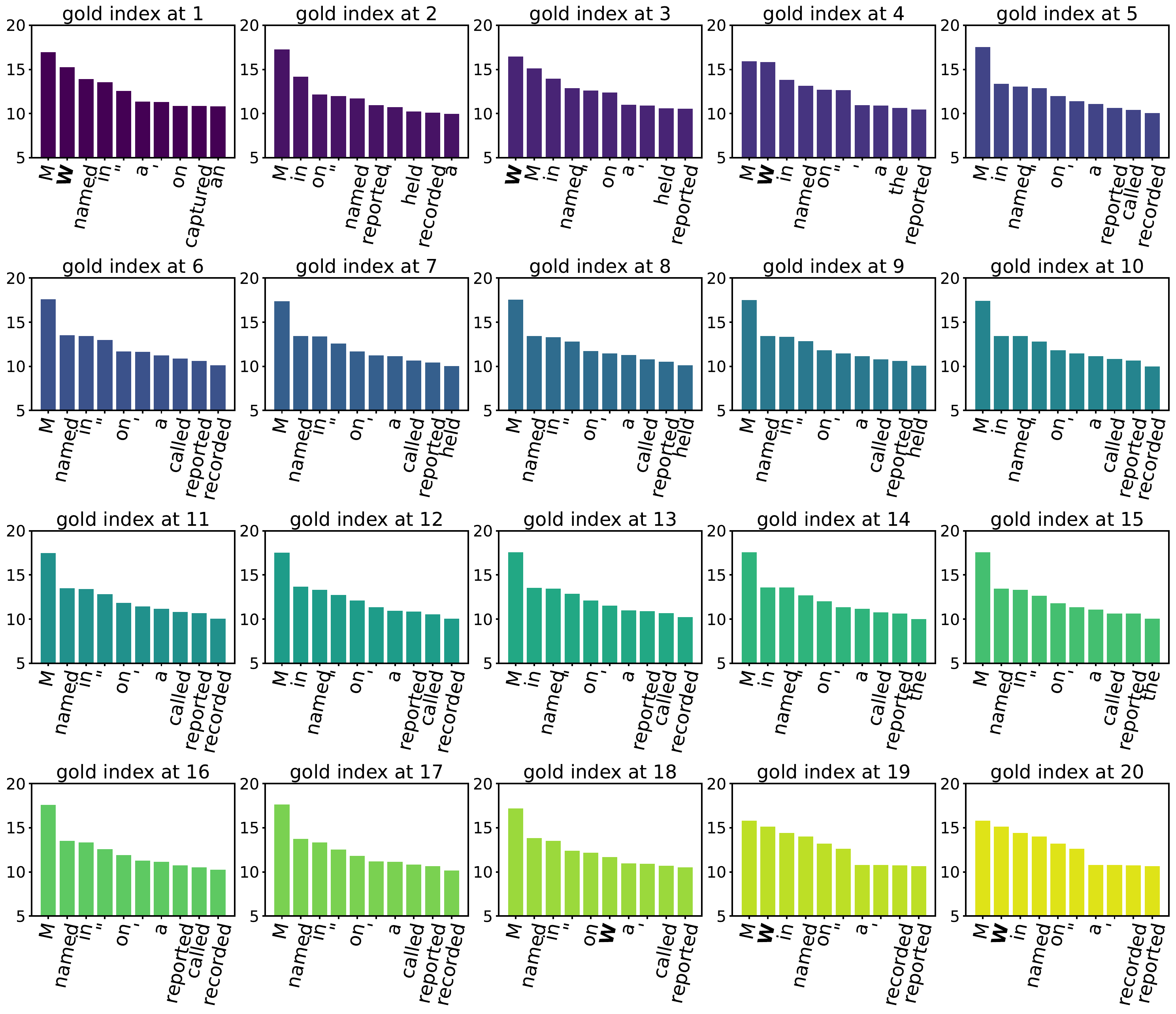}
        \label{fig:Mistral-7B-Instruct-v0.3_logits}
    \end{minipage}
    \vspace{-.22in}
    \caption{Visualization of the top 10 tokens in the logit distributions at the critical divergence points for Meta-Llama-3-8B-Instruct (left) and Mistral-7B-Instruct-v0.3 (right).}
    \label{fig: logits appendix}
\end{figure*}
\vspace{-.1in}
\begin{figure*}[htbp]
    \centering
    \includegraphics[width=0.8\textwidth]{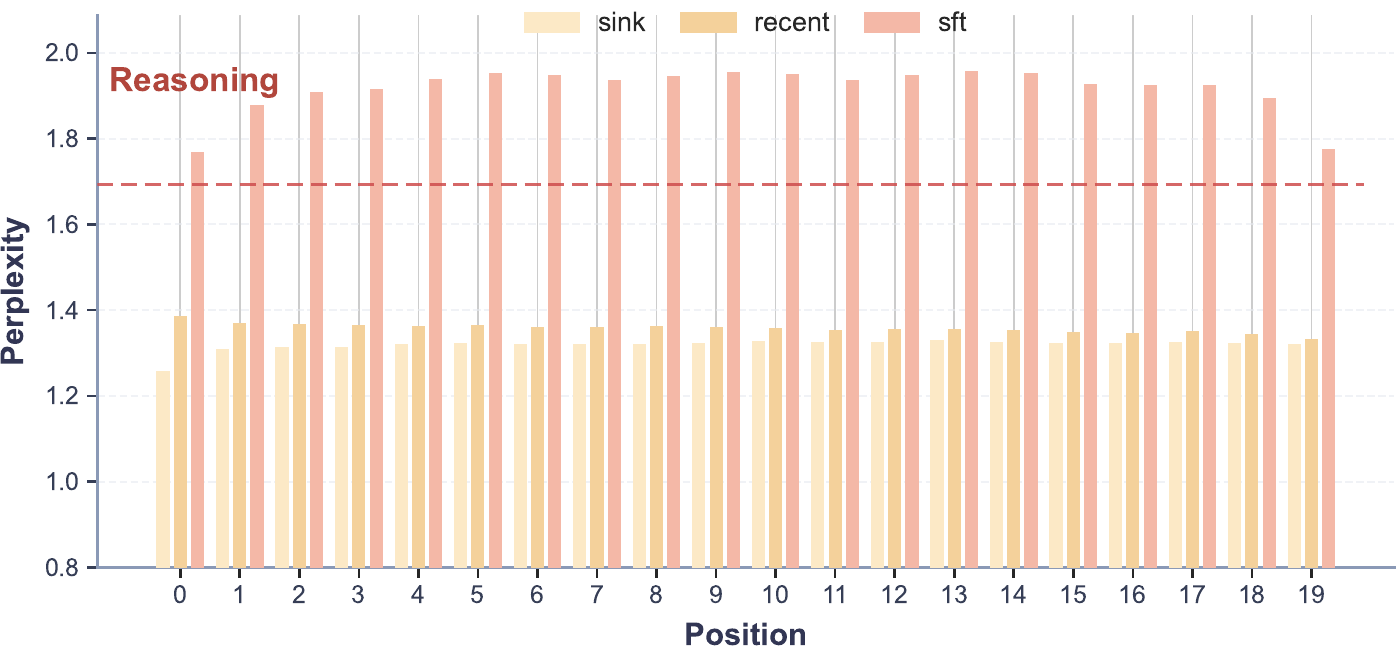}
    \vspace{-.1in}
    \caption{Average PPL over 500 examples for responses collected from sink,recent positions and SFT labels.}
    \label{fig: appendix llama3.1_ppl}
\end{figure*}

\section{Experiments Details.}
\label{appendix: Experiments Details.}
% \paragraph{Data Construction} For retrieval tasks, we set $\mathcal{K}$  as 4 and 
% \paragraph{}
\paragraph{Implementation Details.} 
All experiments are with the following hyperparameters: (1) learning rate of $\alpha=3\times10^{-6}$; (2) a batch size of $b=32$; and (3) $n=2$ training epochs. We employ DeepSpeed ZeRO-3 optimization and FlashAttention~\cite{dao2022flashattention} to accelerate the training process, utilizing the bfloat16 data format. Furthermore, for inference acceleration, we adopt the vLLM ~\cite{kwon2023efficient} framework.
For Retrieval, we conducted training with 300 samples at four distinct locations, each randomly selected from the range of 1 to 20.
For Reasoning, we conducted training on 500 samples. These samples were distributed across four randomly selected pairs of locations, where each individual location within a pair could range from 1 to 20. For both retrieval and reasoning tasks, we only collected a greedy-search response from anchor positions. Commonly, we set the hyperparameter $\lambda$ controlling the intensity of the anchoring loss as 1.0.
All experiments are conducted on NVIDIA H20 GPUs. 

\paragraph{Evaluation Metrics.} For retrieval tasks, we adopt the Sub\_EM metric, while for reasoning tasks, we use the EM metric. Specifically, in reasoning tasks, we first use regular expressions to identify the tokens where the answer appears, and then perform exact match (EM) evaluation on the subsequent part. We test TQA\footnote{https://huggingface.co/datasets/vsearch/tqa}, WebQ \footnote{https://huggingface.co/datasets/vsearch/webq}and NQ \footnote{https://huggingface.co/datasets/vsearch/nq}under retrieval-argumented QA settings and the data sources are from Huggingface.

\paragraph{Baselines.} For Pos2Distill-R\textsuperscript{1}, we introduce vanilla SFT and SeqKD as baselines. SFT directly fine-tunes the student model on data supervised by gold responses, whereas SeqKD fine-tunes the student model on data generated by the teacher model. For Pos2Distill R\textsuperscript{1}, as discussed in related work, mainstream state-of-the-art approaches addressing PB in retrieval tasks include MsPoE ~\cite{zhang2024found}, Attention Buckets ~\cite{chen-etal-2024-fortify}, MoICE~\cite{lin2024mixture}, and PEAR~\cite{yin2024pear}. Although these methods improve performance across positions, PB still persists. All of them are based on mechanistic approaches, and following \cite{yu2024mitigate}, we mainly compare with MsPoE in the main paper. To further demonstrate the effectiveness of our method, we also provide additional comparisons on Llama2-7B-chat-4k~\ref{tab: r1 method_comparison}. Our approach not only better balances performance across positions but also achieves the highest average score of 68.18, which illustrates the advantage compared to previous methods. For Pos2Distill R\textsuperscript{2}, which follows a self-training paradigm using CoT data distilled from advantageous positions, we consider the most recent and relevant baselines, including Longfaith-SFT~\cite{yang2025longfaithenhancinglongcontextreasoning}, longfaith-DPO~\cite{yang2025longfaithenhancinglongcontextreasoning}, and Sealong~\cite{li2024large}. These works involves how to generate high-quality and faithful Chain-of-Thought data for self-improvement and achieving good performance on long-context reasoning tasks. Therefore, we adopt these methods as baselines against to Pos2Distill R\textsuperscript{2} in our experiments.

\paragraph{Prompt Template}
The prompt template can be found in Tab.~\ref{tab:prompt_template}. The Retrieval prompt template instructs the LLM to provide a high-quantify answer (likely meaning a high-quality, precise, or well-supported answer) by exclusively using information from provided  documents, explicitly noting that some documents might be irrelevant. This template emphasizes factual accuracy and direct extraction from given sources, limiting the model's ability to introduce external knowledge.
The Reasoning prompt template outlines a structured, multi-step approach. It first directs the LLM to identify the relevant information from a long context. Next, it requires "step-by-step reasoning based on that information." Finally, it specifies that "The final answer must end with: ‘The answer is:’". This template promotes logical deduction, information synthesis, and a clear, conclusive output format~\cite{yang2025beyond}.
\begin{table}[!h]
\centering
\small
\setlength{\tabcolsep}{1pt} % 调整列间距
\begin{tabular}{lccccccc}
\toprule
\textbf{Method} & \textbf{1} & \textbf{3} & \textbf{5} & \textbf{7} & \textbf{10} & \textbf{Avg.} $\uparrow$ & \textbf{GAP} $\downarrow$ \\
\midrule
base                   & 64.14 & 65.95 & 64.97 & 62.67 & 67.53 & 65.05 & 4.86  \\
+ Ms-PoE              & 66.06 & 64.29 & 63.99 & 62.22 & 64.75 & 64.34 & 3.84  \\
+ AB   & 66.36 & 66.14 & 65.25 & 63.20 & 64.93 & 65.18 & 3.16  \\
+ MoICE               & 65.50 & 66.33 & 65.61 & 64.11 & 65.84 & 65.48 & 2.22  \\
+ PEAR                & 62.71 & 67.01 & 68.32 & 66.44 & 69.57 & 66.81 & 6.86  \\
\rowcolor[rgb]{ .867, .922, .969}+ PosDistill R1 & \textbf{67.27} & \textbf{68.46} & \textbf{69.06} & \textbf{68.66} & \textbf{67.47} & \textbf{68.18} & \textbf{1.79} \\
\bottomrule
\end{tabular}
\caption{Comparison of mainstream state-of-the-art approaches addressing PB in retrieval tasks.}
\label{tab: r1 method_comparison}
\end{table}

\begin{table*}[ht!]
\centering
\resizebox{\linewidth}{!}{
\begin{tabular}{>{\centering\arraybackslash}m{2.5cm} p{16cm}}
\toprule
Category & Prompt Template \\
\midrule
Retrieval &
Please write a high-quantify answer for the given question using only the provided search documents (some of which might be irrelevant).
\\
\midrule
Reasoning & Let’s first identify the relevant information from the long context and list it. Then, carry out step-by-step reasoning based on that information, and finally, provide the answer. The final answer must end with “The answer is:”. \\

\bottomrule
\end{tabular}
}
\caption{Prompt Templates.}
\label{tab:prompt_template}
\end{table*}

\begin{table*}[!t]
\centering
\small
\renewcommand\arraystretch{1}
\setlength{\tabcolsep}{1.8pt}
\resizebox{0.9\linewidth}{!}{
    \begin{tabular}{lccccc|ccccc|ccccc|c}
    \toprule
    &\multicolumn{5}{c|}{\textbf{Connected}}&\multicolumn{5}{c|}{\textbf{Disconnected}}&\multicolumn{5}{c|}{\textbf{Reversed}}\\
    \cmidrule(r){2-17}
    % \cmidrule(r){2-6} \cmidrule(r){7-11} \cmidrule(r){12-16}
    \textsc{Positions}&[0,1]&[5,6]&[12,13]&[17,18]&\textbf{Avg.$\uparrow$ }&[0,8]&[5,13]&[6,14]&[8,16]&\textbf{Avg.$\uparrow$ }&[8,0]&[13,5]&[14,6]&[16,8]&\textbf{Avg.}&\textbf{\textsc{Gap.}	$\downarrow$  } \\
    \midrule
    \textsc{Qwen2.5-3B}&31.2&27.5&28.1&30.2&\textbf{29.3}&28.7&24.3&22.2&23.4&\textbf{24.7}&25.5&24.9&22.6&20.9&\textbf{23.5}&10.3\\
    \textsc{Qwen2.5-14B}&56.6&54.1&54.7&59.5&\textbf{56.2}&55.0&49.5&50.0&51.8&\textbf{51.6}&57.7&51.4&52.0&51.9&\textbf{53.3}&10.0 \\
    \textsc{Qwen2.5-32B}&61.7&59.8&59.1&63.2&\textbf{61.0}&59.4&54.7&54.2&54.7&\textbf{55.8}&60.3&56.6&55.4&57.8&\textbf{57.5}&9.0\\
    \bottomrule

    \end{tabular}}
    \caption{PB in reasoning tasks.}
    \label{tab: pb for reasoning}
    \vspace{-.2in}
\end{table*}

\section{Behavior of PB}
\label{appendix:behavior_position_bias}

For retrieval tasks, we statistically compute the perplexity (PPL) of each response sampled from both the sink and recent positions, conditioned on prompts corresponding to trivial positions (see Fig.~\ref{fig: appendix llama3.1_ppl}). Interestingly, the PPL is much lower compared to that of gold labels from SFT, which strongly indicates a high degree of similarity in the response space when conditioned on different permutations of the same set of documents. A closer examination reveals that LLMs primarily diverge at a few decisive tokens, a phenomenon we refer to as token-shifting. At these key generation steps, as shown in the logit distributions in Fig.~\ref{fig: logits appendix}, the correct token still appears among the top-10 predictions. This observation suggests that LLMs possess the potential to recover from initially incorrect decoding paths, and that responses from the sink position serve as natural gold signals for correction.
In contrast, the dynamics in reasoning tasks are more complex. As shown by the PPL values in Fig.~\ref{fig: appendix llama3.1_ppl}, perplexity increases substantially, indicating that token-shifting is not an appropriate assumption for reasoning scenarios.
\section{Additional Results}
\paragraph{LCLMs with More Documents} Under longer-context settings, we conducted additional experiments with Mistral-7B-Instruct trained on 20 and 50-document contexts respectively, and evaluated them on context lengths ranging from 20 ($\approx$4k tokens) to 80 documents ($\approx$14k tokens), as indicated in Tab. \ref{tab: additional results under more docs}. \textbf{Takeaways:} While training on longer contexts helps further improve the model’s retrieval performance, training with relatively shorter contexts can also yield competitive results. Trained on 50 documents, the model not only generalizes well to fewer documents (achieving high and stable performance, e.g., 77\% accuracy on 20 docs), but also generalizes effectively to more documents (maintaining 70\% accuracy on 80 docs). In addition, models trained on 20 documents still achieve 67\% accuracy when evaluated on 50 docs, exhibiting minimal variance across positions and no significant drop compared to the 69.5\% accuracy observed when tested on 20 docs. This strong out-of-domain generalization, even when trained with fewer docs (e.g., 20), further demonstrates the superiority of our approach.

\section{Statement on AI Usage} 
In the preparation of this manuscript, ChatGPT by OpenAI was utilized solely for the purpose of language refinement and stylistic enhancement. All scientific ideas, methodologies, analyses, and conclusions presented in this work are entirely the authors' own and were developed independently without reliance on AI-generated content. 
\begin{table}[h]
\centering
\small
\setlength{\tabcolsep}{2pt} % 调整列间距
\begin{tabular}{lp{1cm}cccccc}
\toprule
\textbf{Test.} & \textbf{Train.} & \textbf{0\%} & \textbf{25\%} & \textbf{50\%} & \textbf{75\%} & \textbf{100\%} & \textbf{AVG. $\uparrow$} \\
\midrule
20docs  & 20 & 72.3 & 69.5 & 67.5 & 68.5 & 69.7 & 69.5 \\
        & 50 & \textbf{77.8} & 76.2 & 76.6 & 77.2 & 77.4 & \textbf{77.0} \\
30docs  & 20 & 70.3 & 67.9 & 69.3 & 67.3 & 70.9 & 69.1 \\
        & 50 & 74.9 & 75.4 & 75.1 & 76.2 & 77.2 & \textbf{75.8} \\
40docs  & 20 & 67.1 & 66.7 & 68.3 & 66.5 & 68.3 & 67.4 \\
        & 50 & 74.1 & 75.2 & 76.2 & 77.1 & 77.6 & \textbf{75.8} \\
50docs  & 20 & 69.1 & 66.1 & 66.7 & 66.1 & 67.2 & 67.0 \\
        & 50 & 73.5 & 73.9 & 74.5 & 75.2 & 74.9 & \textbf{74.4} \\
70docs  & 20 & 64.9 & 63.9 & 63.9 & 65.1 & 65.1 & 64.6 \\
        & 50 & 69.3 & 69.1 & 71.7 & 70.1 & 70.1 & \textbf{70.1} \\
80docs  & 20 & 64.9 & 63.9 & 63.9 & 65.1 & 65.1 & 64.6 \\
        & 50 & 68.1 & 68.3 & 70.1 & 70.9 & 71.9 & \textbf{69.9} \\
\bottomrule
\end{tabular}
\caption{Performance comparison on different document numbers. The column \textbf{AVG. $\uparrow$} shows the average score across percentages. Bold values indicate the best result in each group.}
\label{tab: additional results under more docs}
\end{table}
% \label{sec:appendix}

\end{document}